\documentclass[10pt,twocolumn,letterpaper]{article}
\usepackage{cvpr}      

\usepackage{tabularray}
\usepackage{multirow}
\usepackage{pifont}
\usepackage[accsupp]{axessibility}
\usepackage{listings}
\usepackage{xcolor}
\usepackage{color}
\definecolor{codegreen}{rgb}{0,0.6,0}
\definecolor{codegray}{rgb}{0.5,0.5,0.5}
\definecolor{codepurple}{rgb}{0.58,0,0.82}
\definecolor{backcolour}{rgb}{0.95,0.95,0.92}
\lstdefinestyle{mystyle}{
    backgroundcolor=\color{backcolour},   
    commentstyle=\color{codegreen},
    keywordstyle=\color{magenta},
    numberstyle=\tiny\color{codegray},
    stringstyle=\color{codepurple},
    basicstyle=\ttfamily\scriptsize,
    breakatwhitespace=false,         
    breaklines=true,                 
    captionpos=b,                    
    keepspaces=true,                 
    numbers=left,                    
    numbersep=5pt,                  
    showspaces=false,                
    showstringspaces=false,
    showtabs=false,                  
    tabsize=2,
    showlines=true
}
\definecolor{cvprblue}{rgb}{0.21,0.49,0.74}
\usepackage[pagebackref,breaklinks,colorlinks,allcolors=cvprblue]{hyperref}
\def\paperID{31651} 
\def\confName{CVPR}
\def\confYear{2026}

\title{Tell2Adapt: A Unified Framework for Source Free Unsupervised Domain Adaptation via Vision Foundation Model}

\author{Yulong Shi$^1$ \quad Shijie Li$^1$ \quad Ziyi Li$^1$ \quad Lin Qi$^{1,2,3}$\thanks{Corresponding author (qilin@bmie.neu.edu.cn).}\thanks{This work was supported by the Key Research and Development Program of Liaoning Province (2024JH2/102500076) and Fundamental Research Funds for the Central Universities (N25BJD013).} \\
$^1$College of Medicine and Biological Information Engineering, Northeastern University \\%
$^2$Engineering Research Center of Medical Imaging and Intelligent Analysis, Ministry of Education\\
$^3$Key Laboratory of Medical Image Computing, Ministry of Education, Northeastern University
}

\begin{document}
\maketitle
\begin{abstract}
Source Free Unsupervised Domain Adaptation (SFUDA) is critical for deploying deep learning models across diverse clinical settings. However, existing methods are typically designed for low-gap, specific domain shifts and cannot generalize into a unified, multi-modalities, and multi-target framework, which presents a major barrier to real-world application. To overcome this issue, we introduce Tell2Adapt, a novel SFUDA framework that harnesses the vast, generalizable knowledge of the Vision Foundation Model (VFM). Our approach ensures high-fidelity VFM prompts through Context-Aware Prompts Regularization (CAPR), which robustly translates varied text prompts into canonical instructions. This enables the generation of high-quality pseudo-labels for efficiently adapting the lightweight student model to target domain. To guarantee clinical reliability, the framework incorporates Visual Plausibility Refinement (VPR), which leverages the VFM's anatomical knowledge to re-ground the adapted model's predictions in target image's low-level visual features, effectively removing noise and false positives. We conduct one of the most extensive SFUDA evaluations to date, validating our framework across 10 domain adaptation directions and 22 anatomical targets, including brain, cardiac, polyp, and abdominal targets. Our results demonstrate that Tell2Adapt consistently outperforms existing approaches, achieving SOTA for a unified SFUDA framework in medical image segmentation. Code are avaliable at \href{https://github.com/derekshiii/Tell2Adapt}{Github}.
\end{abstract} 
\section{Introduction}
\label{sec:Introduction}
\begin{figure}[t]
  \centering
 \includegraphics[width=\linewidth]{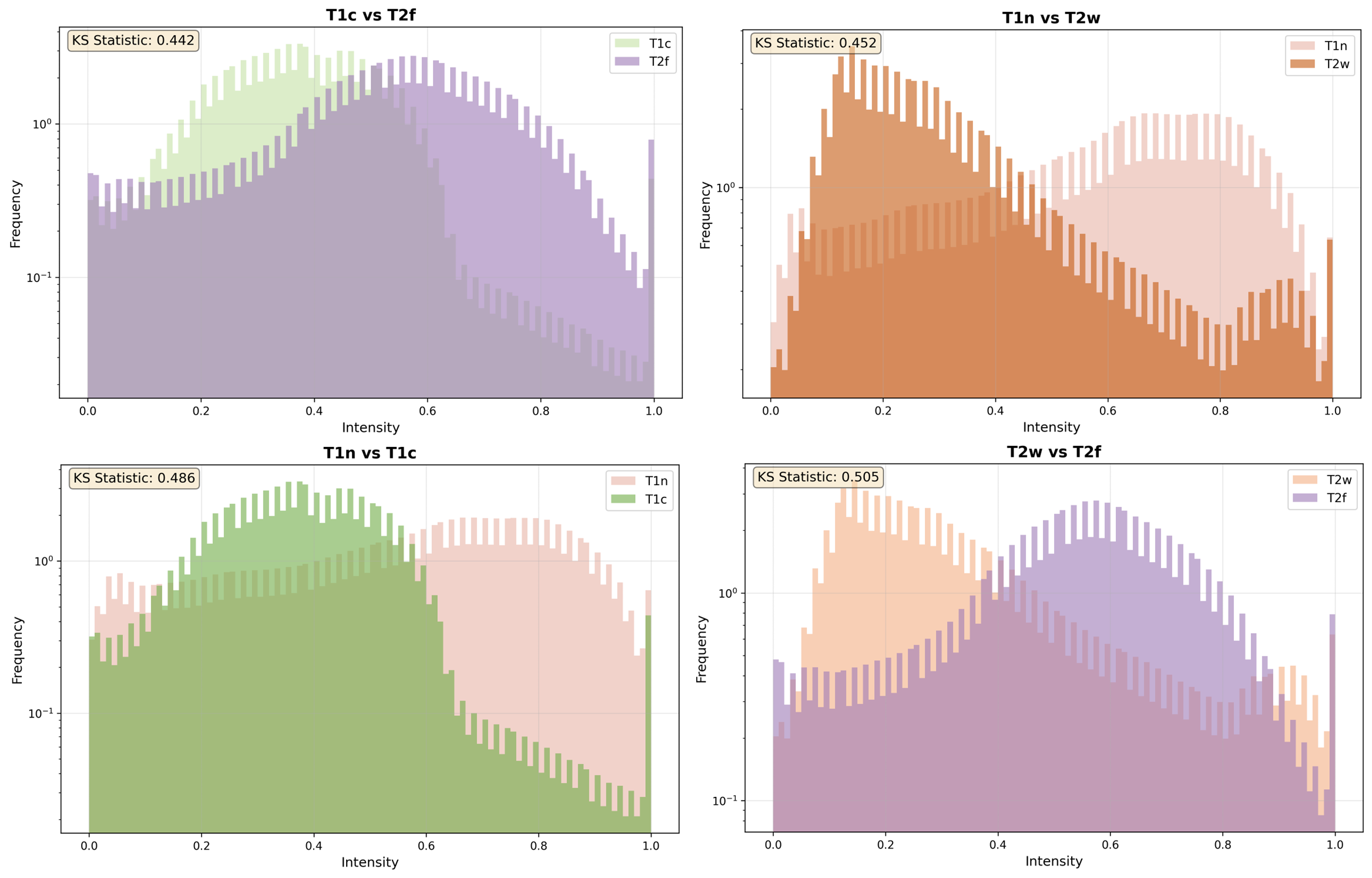}
   \caption{Distribution shift across imaging modalities in BraTS. Kolmogorov-Smirnov statistics quantify the divergence between intensity distributions of the four different imaging modalities.}
   \label{fig:domian_shift}
\end{figure}

Deep learning has demonstrated remarkable success in natural image analysis tasks, yet translating this success to medical image analysis remains uniquely challenging. Unlike natural images, medical images present distinct characteristics, like low spatial resolution, complex tissue backgrounds, and modality-specific artifacts. These intrinsic properties exacerbate the vulnerability of deep learning models to domain shift caused by different scanning devices or imaging modalities \cite{SFUDA_review}. As illustrated in \cref{fig:domian_shift}, the BraTS 2024 (BraTS) dataset shows substantial distribution shifts across different imaging modalities, which often lead to severe performance degradation and pose major barriers to clinical deployment. To overcome these challenges, unsupervised domain adaptation has been developed, but it typically requires concurrent access to source data, which is often infeasible in clinical practice due to patient privacy regulations and data confidentiality agreements \cite{fvp}.

This has given rise to a more practical and challenging paradigm: source free unsupervised domain adaptation, where domain adaptation to an unseen target domain is achieved without any access to the source data \cite{FSM_SFUDA,heal}. The source free and unsupervised settings are particularly appealing in clinical scenarios, in which data privacy is paramount. Despite significant progress, current SFUDA methods suffer from a critical limitation: most existing works focus on low-gap adaptation and specific anatomical targets \cite{ttsfuda} \cite{wsmsfuda} \cite{Zeng2024ReliableSA}. These specialized methods lack the flexibility and scalability to function as a unified solution for the wide variety of modalities and anatomical targets encountered in clinical practice. Consequently, a unified SFUDA framework remains a significant challenge.

Recently, the advent of VFMs has marked a paradigm shift in Medical image analysis. General VFMs like DINO \cite{DINO}, CLIP \cite{CLIP}, and SAM \cite{SAM}, and medical VFMs like MedSAM \cite{MedSAM}, MedSAM-3D\cite{medsam3d}, VividMed \cite{vividmed}, BiomedParse \cite{biomedparse}, and BiomedCLIP \cite{BiomedCLIP} have demonstrated strong generalization capabilities and performance. For SFUDA, these models as an ideal, untapped external knowledge source to generate supervisory signals for the target domain. Guided by spatial or text prompts, VFM offers a transformative paradigm that bypasses the reliance on source data and establishes a foundation for a unified SFUDA framework across multi-modalities and multi-target medical image segmentation.
\begin{figure}[t]
  \centering
 \includegraphics[width=1.05\linewidth]{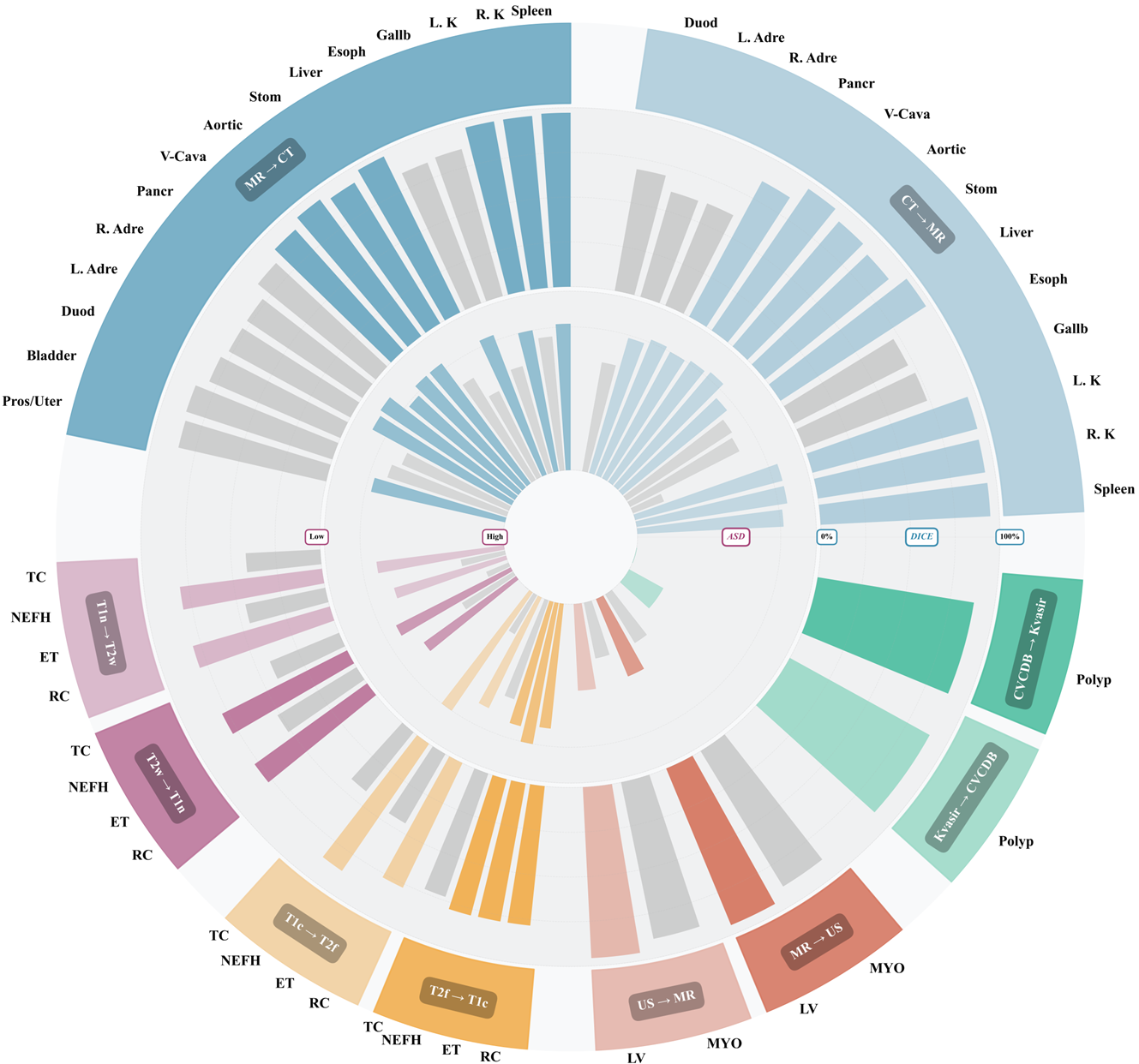}
   \caption{Generalization performance of Tell2Adapt across diverse adaptation tasks. The outermost ring displays the adaptation direction, the middle ring shows Dice score (DICE), while the innermost ring represents the Average Surface Distance (ASD).}
   \label{fig:test_data}
\end{figure}

In this work, we propose Tell2Adapt, a unified SFUDA framework that achieves generalizable domain adaptation across an unprecedented range of modalities and targets, as illustrated in \cref{fig:test_data}. Specifically, we first introduce Context-Aware Prompts Regularization that employs a Large Language Model (LLM) to interpret and standardize arbitrary user text prompts into a canonical instruction set, ensuring reliable guidance for VFM to generate high-quality pseudo-labels for the unlabeled target domain. Subsequently, this knowledge is efficiently distilled into the lightweight source model. This knowledge transfer not only imbues the source model with rich, domain-specific information but also compresses the VFM's generalist knowledge into a compact, specialist model. Finally, we incorporate Visual Plausibility Refinement that verifies the adapted model's prediction against the original target image. This step filters out anatomically implausible regions by ensuring their low-level visual features are consistent with the general anatomical properties expected for that category, effectively re-grounding the prediction in the target image. Our main contributions are threefold: 
\begin{enumerate}[noitemsep]
    \item We propose Tell2Adapt, a unified SFUDA framework for medical image segmentation that harnesses multi-modalities knowledge from VFM. The framework's effectiveness as a unified solution is demonstrated through the most extensive and challenging SFUDA evaluations to date, achieving SOTA across 10 domain adaptation directions and 22 distinct anatomical targets, diverse from abdominal, cardiac, polyp, and brain targets.
    \item We introduce Context-Aware Prompts Regularization, which parses the internal context of varied inputs and transforms them into a canonical, structured format. By resolving ambiguity and enriching the prompts with essential contextual information, this module provides the VFM with standardized and reliable guidance. 
    \item We propose Visual Plausibility Refinement, which significantly enhances the anatomical plausibility and reliability of the predictions. Leveraging the VFM's anatomical knowledge to re-ground the adapted model's predictions in the target image's low-level features, VPR effectively removes noise and false positives in predictions.
\end{enumerate}
\section{Related Works}
\subsection{Source Free Unsupervised Domain Adaptation}
SFUDA aims to adapt a model pre-trained on a source domain to an unseen target domain without accessing the source data, which is critically important due to growing concerns over data privacy and the computational burden of storing and reusing large source datasets. Mainstream SFUDA primarily falls into three categories: data synthesis, knowledge distillation, and self-training methods. Data synthesis methods typically generate source-like data that helps bridge the domain shifts \cite{Zeng2024ReliableSA}. Knowledge distillation methods employ a teacher-student strategy to improve pseudo-label quality and enhance training stability \cite{CHEN2024182}. Self-training methods generate pseudo-labels on the target domain for prototype alignment and encourage adaptation by contrast learning  \cite{protocontra} or denoising \cite{A3DualUD}.  Despite their progress, a fundamental limitation plagues the vast majority of existing methods: they are typically designed for low-gap and specific target adaptation. There is currently no unified SFUDA framework capable of handling a wide array of diverse, multi-target, and huge-gap domain adaptation with a single, consistent methodology.
\subsection{Vision Foundation Models in Medical Scenario}
\begin{figure*}[ht]
  \centering
 \includegraphics[width=\linewidth]{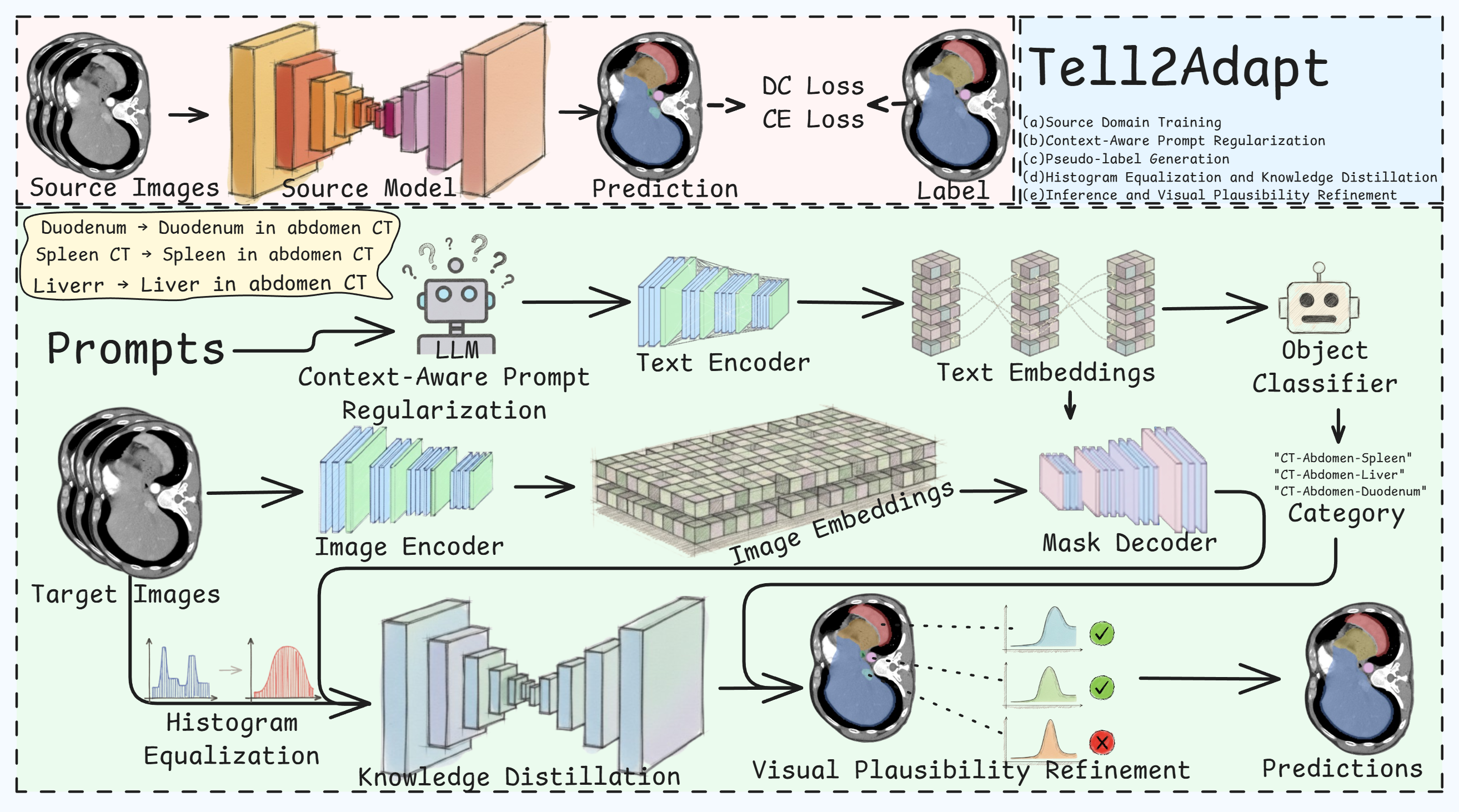}
   \caption{Overview of Tell2Adapt, a unified SFUDA framework. The workflow decouples pseudo-label generation from the source model by leveraging VFM guided by text prompts. We introduce CAPR to standardize these prompts, enabling the VFM to generate high-fidelity pseudo-labels. This knowledge is then distilled into the lightweight source model via self-training and refined using VPR.}
   \label{fig:WholePlot}
\end{figure*}
The advent of VFMs has recently catalyzed a paradigm shift in medical image analysis. These emerging models can be broadly categorized by their prompting mechanism: spatial-prompted and text-prompted. The spatial-prompted models include general-purpose VFMs like SAM, DINO, and their medically-adapted derivatives (MedSAM, MedSAM-3D), which are fine-tuned on large medical datasets. While powerful, these models are unsuitable for SFUDA due to the reliance on high-quality spatial prompts. In the source-free setting, spatial prompts are generated from the source model’s predictions, which are often unreliable under severe domain shift. Using these noisy predictions as prompts triggers an error propagation process, causing the VFM to produce highly inaccurate pseudo-labels.

Recognizing this limitation, text-prompted VFMs have been developed, often from the ground up on specialized medical data, integrating language as a more powerful and flexible prompting modality. VividMed and BiomedParse fall into this category, with BiomedParse representing the SOTA in this direction by enabling segmentation, detection, and recognition guided purely by text prompts. This ability to interpret high-level semantic instructions without requiring precise spatial prompts makes it an ideal source of external knowledge, breaking the cycle of error accumulation and perfectly suiting the challenges of SFUDA.
\subsection{Synergizing VFM with SFUDA}
The rich and generalizable knowledge encapsulated within VFM presents a transformative opportunity for SFUDA. The core idea is using VFM to provide high-quality supervision for the unseen target domain, thereby bypassing the limitations of traditional SFUDA methods that rely solely on the unreliable pseudo-labels of the source model. Recent pioneering works have begun to explore this direction. For example, some studies have utilized MedSAM to generate pseudo-labels \cite{SRPL} \cite{iplc} \cite{iplc+} or to guide the training process \cite{DFG} for domain adaptation. These initial efforts have validated the potential of the combination of VFM and SFUDA. However, these works rely on the source model's low-quality predictions as spatial prompts, which creates a potential failure cascade due to a large domain shift.
\begin{table}[t]
\centering
\small
\caption{Comparison of prompts before and after regularization}
\label{tab:prompt_comparison}
\begin{tabular}{p{0.33\linewidth}|p{0.5\linewidth}}
\hline
\hline
\textbf{Original Prompts} & \textbf{Regularized Prompts} \\
\hline
Liverr [SEP] Spleen CT [SEP] Abdomen duodenum & Liver in abdomen CT [SEP] Spleen in abdomen CT [SEP] Duodenum in abdomen CT \\
\hline
Right ventricle [SEP] Myocardium MRI [SEP] Left ventricle & Right ventricle in cardiac MRI [SEP] Myocardium in cardiac MRI [SEP]  Left ventricle in cardiac MRI \\
\hline
\hline

\end{tabular}
\end{table}
\section{Methods}
Our proposed unified SFUDA framework, Tell2Adapt, is designed to distill the generalizable knowledge of a VFM into the lightweight source model, making it practical for clinical deployment. At its core, the framework leverages BiomedParse, guided by canonical text prompts regularized by CAPR to generate high-quality pseudo-labels for the target domain. This rich supervisory signal then drives the adaptation of the source model through a knowledge distillation paradigm. Finally, during inference, we introduce VPR to refine the final predictions from the adapted model, ensuring anatomical plausibility. Formally, we define the source domain as $D_S = \{x_i^s, y_i^s\}_{i=1}^{N_s}$ and the unlabeled target domain as $D_T = \{x_j^t\}_{j=1}^{N_t}$. In SFUDA settings, only the model $M_s$ pre-trained on $D_S$ and the target images $D_T$ are available during the adaptation process.
\subsection{Context-Aware Prompts Regularization}
\label{sec:CAPR}
Our framework employs a VFM guided by text prompts to generate pseudo-labels for the target domain, while its performance remains highly sensitive to the syntactic and semantic structure of the prompts. In real-world clinical applications, prompts can be noisy, ambiguous. Furthermore, a single query often contains multiple, semantically-related targets, which provides rich but unstructured intra-prompt context. This variability can introduce semantic drift in the VFM's text encoder, leading to inconsistent embeddings and erroneous pseudo-labels. To address this critical challenge, we introduce CAPR, which acts as a semantic normalizer to canonicalize the prompts. Unlike simple prompt cleaning, CAPR performs a holistic analysis of the entire string. We leverage the powerful in-context reasoning and instruction following capabilities of LLM.

The LLM performs global context inference by following a meta-prompt that instructs it to collectively analyze all input prompts and identify a shared, high-level context, such as the imaging modality and anatomical region. Second, using this inferred global context, the LLM conducts contextual enrichment and canonicalization for each sub-prompt. It corrects typographical errors, resolves ambiguities, and enriches the sub-prompt with the shared context. As shown in \cref{tab:prompt_comparison}, each sub-prompt is then reformatted into a predefined, descriptive, and consistent structure, which we define as the canonical format: [Target] in [Anatomical Site] [Modality]. The structured strings returned by the LLM are then parsed to serve as clean, canonical prompts for the VFM's text encoder. 
\subsection{VFM-Guided Knowledge Distillation}
With the canonical instructions from CAPR, Tell2Adapt leverages BiomedParse as an external knowledge source to generate supervisory signals for the unlabeled target domain $D_T$. This process is followed by a knowledge distillation paradigm that distills this knowledge into the lightweight and deployable source model.

The initial step is to generate high-quality pseudo-labels for the target images. BiomedParse comprises a text encoder, an image encoder, a mask decoder, and an object classifier, all jointly trained to understand visual-semantic relationships. For each target image $x^t \in D_T$, the regularized text prompt is converted into text embeddings by the text encoder, while the target image is transformed into rich spatial feature embeddings by the image encoder. The mask decoder then performs a grounding operation, using the text embeddings to query the image embeddings and attend to the relevant anatomical regions. This yields segmentations that serve as pseudo-label $\hat{y}^t$ for the image $x^t$.

After generating the pseudo-label set $\{\hat{y}_j^t\}_{j=1}^{N_t}$, we address the critical challenge of low-level domain shift. As visualized in \cref{fig:domian_shift}, variations in intensity distributions are a primary barrier to adaptation. To mitigate this and create more robust training signals, we apply standard histogram equalization (HE) to target images $x_j^t$. This step normalizes global image contrast by redistributing pixel intensities based on their cumulative distribution function, compelling the model to focus on learning invariant anatomical features rather than superficial domain-specific statistics.

The knowledge distillation then proceeds by training the source model on the enhanced target images and their corresponding pseudo-labels, forming the training set: \begin{equation}\mathcal{D}_T^{adapt} = \{(\text{HE}(x_j^t), \hat{y}_j^t)\}_{j=1}^{N_t}\end{equation} The model is optimized by minimizing a combination of Dice and Cross-Entropy losses. This procedure effectively distills generalized knowledge from BiomedParse into an efficient student model, adapting it to the target domain. The final result is a lightweight, high-performance model that can be deployed clinically without requiring the resource-intensive VFM at inference. Specifically, while direct inference on the BiomedParse requires 371.8M parameters and a peak VRAM of 26.1\,GB on abdominal datasets, our adapted model achieves robust performance using only 31.1M parameters and 4.8\,GB of peak VRAM. This massive reduction in computational overhead significantly lowers the hardware barrier for real-world clinical deployment.
\subsection{Visual Plausibility Refinement}
\begin{figure*}[ht]
  \centering
 \includegraphics[width=\linewidth]{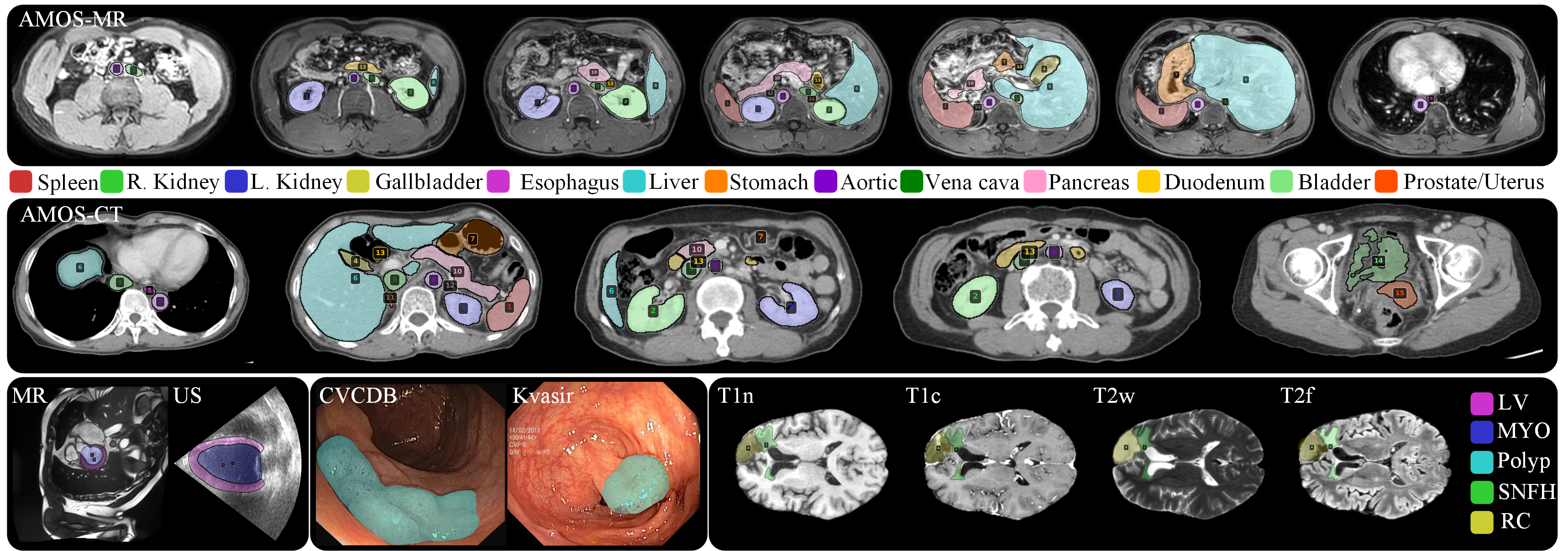}
   \caption{Qualitative results of Tell2Adapt demonstrate its strong generalization capability across diverse and challenging SFUDA directions. The figure presents segmentation outputs from Tell2Adapt adapted to various target domains. Specifically, the first and second rows show MR-CT adaptation for abdominal targets in AMOS, while the third row illustrates extreme-gap adaptation in cardiac targets (MR-US), single target adaptation in polyp (Kvasir-CVCDB), and multi-sequence adaptation in brain targets (T1n-T2w and T1c-T2f).}
   \label{fig:ExampleResult}
\end{figure*}
While the adapted model provides efficient segmentation, its predictions $P_{initial}$ may occasionally contain anatomically implausible regions and misidentifications due to residual domain shift. To enhance the robustness and clinical reliability of final predictions, we introduce a refinement that leverages a rich knowledge of statistical priors pre-computed from BiomedParse.

This process begins by identifying the target’s general anatomical category $C$ using BiomedParse’s inherent object classifier. We then leverage the anatomical priors provided by BiomedParse, which contain pre-computed Beta distributions describing the visual properties of each object category. For any given category $C$, we retrieve a set of four Beta distributions, denoted as $\mathcal{P}_C$.
\begin{equation}\mathcal{P}_C={(\alpha_k,\beta_k)}_{k=1}^4\end{equation}
Each parameter pair models the distribution of a specific visual attribute of the target object: $(\alpha_1, \beta_1)$ represents the distribution of the average object pixel probability, while $(\alpha_2, \beta_2)$, $(\alpha_3, \beta_3)$, and $(\alpha_4, \beta_4)$ collectively model the distributions of the average intensity values in the R, G, and B channels, respectively. For grayscale images, the single channel is duplicated to match the 3-channel format.

The refinement process utilizes these statistical priors to validate and correct the initial predictions $P_{initial}$. Let the initial prediction be a set of $N$ disconnected components $P_{initial} = \{p_1, p_2, \dots, p_N\}$. For each component $p_i$, we compute its corresponding set of four features $\{f_{i,k}\}_{k=1}^4$. We then calculate an anatomical plausibility score, $S(p_i)$, which measures how well each component's features align with the retrieved prior distributions by modeling their joint probability density using Beta distributions:
\begin{equation}S(p_i)=\prod_{k=1}^4\frac{f_{i,k}^{\alpha_k-1}(1-f_{i,k})^{\beta_k-1}}{\mathrm{B}(\alpha_k,\beta_k)}\end{equation}
where $\mathrm{B}(\alpha_k, \beta_k)$ is the Beta function. This score, $S(p_i)$, will be high if the component $p_i$ has a pixel probability and color profile that are statistically typical for the target category $C$, and low otherwise. Finally, the refined prediction, $P_{final}$, is generated by retaining only those components whose plausibility score exceeds a predefined threshold $\tau$:
\begin{equation}P_{final} = \bigcup_{i | S(p_i) >= \tau} p_i\end{equation}
The threshold $\tau$ is defined as two standard deviations $\sigma_S$ below the mean score $\mu_S$ of all $N$ components ($\tau = \mu_S - 2\sigma_S$). This procedure effectively removes noise and false positive regions that are inconsistent with the anatomical knowledge, yielding a more reliable segmentation.
\section{Experiments and Results}
\label{sec:Experiments}
We conduct extensive experiments across 10 domain adaptation directions and 22 anatomical targets, covering abdominal, brain, polyp, and cardiac targets, to comprehensively assess the performance of Tell2Adapt.
\subsection{Dataset and Implementation Details}
\begin{table*}[ht]
\centering
\caption{DICE (\%, mean $\pm$ std) of segmentation results on abdominal targets. ASD for all anatomical targets is provided in Appendix.} 
\label{tab:AMOS}

\resizebox{2\columnwidth}{!}{%
\begin{tabular}{@{}ccccccccccccccccccc@{}}
\toprule
\multicolumn{17}{c}{\textbf{MR$\rightarrow$CT}} \\ \midrule

{ \textbf{Methods}} & { \textbf{VFM}} & { \textbf{Spleen}} & { \textbf{R. K}} & { \textbf{L. K}} & { \textbf{Gallb}} & { \textbf{Esoph}} & { \textbf{Liver}} & { \textbf{Stom}} & { \textbf{Aortic}} & { \textbf{V-Cava}} & { \textbf{Pancr}} & { \textbf{R. Adre}} & { \textbf{L. Adre}} & { \textbf{Duod}} & { \textbf{Bladder}} & { \textbf{Pros/Uter}} & { \textbf{mDICE}} \\ \midrule

{ Baseline} & $\times$ & { 65.1$\pm$30.4} & { 71.2$\pm$26.9} & { 63.2$\pm$29.6} & { 27.2$\pm$27.7} & {42.2$\pm$22.1} & {82.2$\pm$9.9} & {46.9$\pm$28.6} & {72.0$\pm$20.3} & {61.7$\pm$23.7} & {55.7$\pm$23.3} & {39.7$\pm$22.8} & {44.7$\pm$24.2} & {38.5$\pm$22.4} & {0.0$\pm$0.0} & {0.0$\pm$0.0} & {47.4$\pm$24.7} \\

{ Supervised} & $\times$ & {97.2$\pm$1.7} & {96.4$\pm$1.7} & {94.7$\pm$11.7} &86.0$\pm$13.6&84.1$\pm$10.0& 97.6$\pm$1.0 & 93.1$\pm$5.1 &95.3$\pm$2.8&90.2$\pm$5.7&83.7$\pm$12.9&77.8$\pm$11.0 &78.9$\pm$11.7 &81.6$\pm$10.9& 83.3$\pm$19.3 &85.5$\pm$7.0 & {88.4$\pm$7.5}\\ \midrule

{ UPL \cite{upl}} & $\times$ & \textcolor{blue}{\textbf{23.9$\pm$2.0}} & {7.7$\pm$0.2} & \textcolor{blue}{\textbf{30.2$\pm$26.9}} & {0.0$\pm$0.0} & {12.5$\pm$12.7} & {23.6$\pm$30.9} & {10.7$\pm$12.2} & {9.3$\pm$10.0} & {0.7$\pm$1.4} & {8.6$\pm$8.3} & {0.0$\pm$0.0} & {0.0$\pm$0.0} & {2.0$\pm$7.2} & {0.0$\pm$0.0} & {0.0$\pm$0.0} & {8.6$\pm$9.4} \\

{ ProtoContra \cite{protocontra}} & $\times$ & {5.6$\pm$6.3} & {11.9$\pm$2.3} & {7.3$\pm$2.6} & \textcolor{blue}{\textbf{4.3$\pm$3.1}} & {37.5$\pm$24.1} &{40.1$\pm$24.8} & {25.1$\pm$21.3} &{38.6$\pm$27.8} & \textcolor{blue}{\textbf{29.2$\pm$18.9}} & {39.1$\pm$20.1} & {0.0$\pm$0.0} & {0.0$\pm$0.0} & {17.6$\pm$16.2} & {0.0$\pm$0.0} & {0.0$\pm$0.0} & {17.1$\pm$15.9} \\

{ IAPC \cite{IAPC}} & $\times$ & {8.7$\pm$4.5} & {4.5$\pm$2.3} & {1.9$\pm$1.4} & {1.3$\pm$1.5} & {4.5$\pm$11.3} & {15.2$\pm$2.1} & {14.3$\pm$5.6} & { 9.7$\pm$5.0} & {4.6$\pm$1.1} & {2.0$\pm$0.6} & {0.0$\pm$0.0} & {0.0$\pm$0.0} & {2.2$\pm$7.4} & \textcolor{blue}{\textbf{19.8$\pm$16.8}} & \textcolor{blue}{\textbf{7.2$\pm$6.8}} & {6.4$\pm$6.1}\\

{ DFG \cite{DFG}} & \checkmark &{0.7$\pm$0.6} & \textcolor{blue}{\textbf{28.7$\pm$17.4}} & {4.1$\pm$7.2} & {0.0$\pm$0.0} & {35.1$\pm$16.3} & {35.0$\pm$19.2} & {16.6$\pm$16.1} & {13.0$\pm$16.1} & {22.0$\pm$23.3} & {29.7$\pm$18.4} & {0.0$\pm$0.0} & {0.0$\pm$0.0} & {21.2$\pm$12.7} & {0.0$\pm$0.0} & {0.0$\pm$0.0} & {13.7$\pm$13.8}\\

{ IPLC \cite{iplc}} & \checkmark & {0.0$\pm$0.0} & {0.0$\pm$0.0} & {0.0$\pm$0.0} & {0.0$\pm$0.0} & {0.0$\pm$0.0} & \textcolor{blue}{\textbf{65.4$\pm$11.3}} & {10.5$\pm$7.1} & {6.6$\pm$8.3} & {4.1$\pm$2.7} & {3.8$\pm$3.7} & {0.0$\pm$0.0} & {0.0$\pm$0.0} & {0.0$\pm$0.0} & {0.0$\pm$0.0} & {0.0$\pm$0.0} & { 6.0$\pm$16.7}\\ 

{ SRPL \cite{SRPL}} & \checkmark & {0.0$\pm$0.0} & {0.0$\pm$0.0} & {0.0$\pm$0.0} & {0.0$\pm$0.0} & \textcolor{blue}{\textbf{41.0$\pm$25.3}} & {47.6$\pm$29.1} & \textcolor{blue}{\textbf{42.2$\pm$26.0}}& \textcolor{blue}{\textbf{66.0$\pm$28.9}} & {26.2$\pm$30.2} & \textcolor{blue}{\textbf{46.6$\pm$23.3}} & {0.0$\pm$0.0} & {0.0$\pm$0.0} & \textcolor{blue}{\textbf{29.4$\pm$17.9}}  & {0.0$\pm$0.0} & {0.0$\pm$0.0} & \textcolor{blue}{\textbf{19.9$\pm$22.9}} \\ \midrule

\textbf{Ours} & \checkmark & \textcolor{red}{\textbf{97.1$\pm$1.6}} & \textcolor{red}{\textbf{96.3$\pm$2.9}} & \textcolor{red}{\textbf{96.2$\pm$4.6}} & \textcolor{red}{\textbf{85.2$\pm$16.4}} & \textcolor{red}{\textbf{84.0$\pm$8.6}} & \textcolor{red}{\textbf{96.9$\pm$2.6}} & \textcolor{red}{\textbf{91.9$\pm$7.5}} & \textcolor{red}{\textbf{94.9$\pm$2.2}} & \textcolor{red}{\textbf{90.5$\pm$5.5}} & \textcolor{red}{\textbf{84.7$\pm$10.5}} & \textcolor{red}{\textbf{77.4$\pm$8.1}} & \textcolor{red}{\textbf{77.3$\pm$10.3}} & \textcolor{red}{\textbf{79.7$\pm$14.2}} & \textcolor{red}{\textbf{85.8$\pm$16.4}} & \textcolor{red}{\textbf{84.6$\pm$9.2}} & \textcolor{red}{\textbf{88.2$\pm$7.1}}\\

\bottomrule
\end{tabular}%
}

\resizebox{2\columnwidth}{!}{%
\begin{tabular}{@{}cccccccccccccccc@{}}
\toprule
\multicolumn{15}{c}{\textbf{CT$\rightarrow$MR}} \\ \midrule

{ \textbf{Methods}} & { \textbf{VFM}} & { \textbf{Spleen}} & { \textbf{R. K}} & { \textbf{L. K}} & { \textbf{Gallb}} & { \textbf{Esoph}} & { \textbf{Liver}} & { \textbf{Stom}} & { \textbf{Aortic}} & { \textbf{V-Cava}} & { \textbf{Pancr}} & { \textbf{R. Adre}} & { \textbf{L. Adre}} & { \textbf{Duod}} & { \textbf{mDICE}}\\ \midrule

{ Baseline} & $\times$ & { 58.6$\pm$47.0} & { 86.6$\pm$9.2} & { 89.6$\pm$0.7} & {53.1$\pm$28.8} & 58.9$\pm$34.2 & {73.4$\pm$33.0} & 60.0$\pm$19.5 & 61.4$\pm$29.8 & 64.0$\pm$25.1 &65.0$\pm$19.2 & 56.5$\pm$3.3 & 64.7$\pm$10.2 & 44.3$\pm$25.0 &  64.3$\pm$12.5\\

{ Supervised}  & $\times$&97.6$\pm$1.5& 97.1$\pm$0.8&97.1$\pm$0.5&88.1$\pm$10.0 & 73.7$\pm$15.0 &98.0$\pm$1.2 & 90.4$\pm$3.3 & 92.9$\pm$5.2 & 91.7$\pm$2.5 & 86.4$\pm$3.9 &67.7$\pm$9.9 & 71.0$\pm$11.2 & 69.1$\pm$8.7 & 86.2$\pm$10.1\\ \midrule

{ UPL \cite{upl}} & $\times$ & {1.7$\pm$5.3} & {0.8$\pm$1.4} & {2.3$\pm$0.9} & {0.0$\pm$0.0} & {0.0$\pm$0.0} & {12.9$\pm$12.5} & {1.0$\pm$1.1} & {38.4$\pm$19.9} & { 0.0$\pm$0.0} & {12.9$\pm$13.5} & {0.0$\pm$0.0} & {0.0$\pm$0.0} & {0.0$\pm$0.0} & {5.4$\pm$10.0}\\

{ ProtoContra \cite{protocontra}} & $\times$ & {17.8$\pm$16.1} & {4.9$\pm$4.9} & {16.1$\pm$17.8} & {0.0$\pm$0.0} & \textcolor{blue}{\textbf{47.7$\pm$28.0}} & {11.7$\pm$14.4} & {18.7$\pm$14.4} & {6.0$\pm$11.8} & \textcolor{blue}{\textbf{38.2$\pm$24.2}} & {30.5$\pm$16.8} & \textcolor{blue}{\textbf{27.3$\pm$22.7}} & {0.0$\pm$0.0} & {0.0$\pm$0.0} & {16.8$\pm$14.5}\\

{ IAPC \cite{IAPC}} & $\times$ & \textcolor{blue}{\textbf{21.9$\pm$3.3}} & {8.9$\pm$3.8} & {19.4$\pm$6.5} & {0.0$\pm$0.0} & {16.1$\pm$6.9} & {13.6$\pm$23.0} & \textcolor{blue}{\textbf{33.8$\pm$3.7}} & {7.8$\pm$1.5} & {2.6$\pm$8.7} & {0.6$\pm$1.0} & {0.0$\pm$0.0} & {0.0$\pm$0.0}  & {0.0$\pm$0.0} & {9.6$\pm$9.9}\\

{ DFG \cite{DFG}} & \checkmark & {14.0$\pm$12.9} & \textcolor{blue}{\textbf{10.2$\pm$2.7}} & {11.4$\pm$14.7} & {0.0$\pm$0.0} & {36.9$\pm$22.5} & {45.0$\pm$26.9} & {5.9$\pm$10.4} & \textcolor{blue}{\textbf{65.3$\pm$23.3}} & {23.1$\pm$17.7} & {27.3$\pm$22.5} & {0.0$\pm$0.0}  & {0.0$\pm$0.0} & {0.0$\pm$0.0} & {18.4$\pm$19.7} \\

{ IPLC \cite{iplc}} & \checkmark & {0.0$\pm$0.0} & {0.0$\pm$0.0} & {0.0$\pm$0.0} & {0.0$\pm$0.0} & {2.2$\pm$1.3} & \textcolor{blue}{\textbf{59.6$\pm$13.0}} & {5.0$\pm$2.8} & {10.6$\pm$5.8} & {4.6$\pm$2.9} & {8.4$\pm$6.5} & {0.0$\pm$0.0} & {0.0$\pm$0.0} & {17.6$\pm$2.0} & {7.1$\pm$15.8}\\

{ SRPL \cite{SRPL}} & \checkmark & {0.0$\pm$0.0} & {0.0$\pm$0.0} & \textcolor{blue}{\textbf{69.8$\pm$24.6}} & {0.0$\pm$0.0} & {0.0$\pm$0.0} & {38.6$\pm$23.9} & {31.1$\pm$26.4} & {54.0$\pm$28.9} & {0.0$\pm$0.0} & \textcolor{blue}{\textbf{42.1$\pm$26.2}} & {0.0$\pm$0.0}& {0.0$\pm$0.0}   & \textcolor{blue}{\textbf{19.7$\pm$17.2}} & \textcolor{blue}{\textbf{19.6$\pm$24.8}}\\ \midrule

{ \textbf{Ours}} & \checkmark & \textcolor{red}{\textbf{94.8$\pm$3.8}} & \textcolor{red}{\textbf{94.4$\pm$0.8}} & \textcolor{red}{\textbf{95.2$\pm$0.8}} & \textcolor{red}{\textbf{73.3$\pm$7.4}} & \textcolor{red}{\textbf{71.6$\pm$8.9}} & \textcolor{red}{\textbf{96.8$\pm$0.4}} & \textcolor{red}{\textbf{88.9$\pm$7.2}} & \textcolor{red}{\textbf{91.5$\pm$2.7}} & \textcolor{red}{\textbf{85.7$\pm$4.3}} & \textcolor{red}{\textbf{85.9$\pm$5.7}} & \textcolor{red}{\textbf{61.5$\pm$3.0}} & \textcolor{red}{\textbf{61.0$\pm$14.8}} & \textcolor{red}{\textbf{68.9$\pm$6.3}} & \textcolor{red}{\textbf{82.3$\pm$12.9}}\\ 
\bottomrule
\end{tabular}%
}

\footnotesize{~~~~\textit{Note}:  A DICE of 0.0 indicates a complete prediction failure for the corresponding targets.}
\vspace{-3mm}
\end{table*}
\begin{table*}[tbp]
\centering
\caption{DICE and ASD of segmentation results on brain targets.}
\label{tab:BraTS}
\resizebox{2\columnwidth}{!}{%
\renewcommand{\arraystretch}{0.9}  
\begin{tabular}{@{}c@{\hspace{0.3cm}}c cccccccc cccccccc@{}}
\toprule
\multirow{3}{*}{\textbf{Methods}} & \multirow{3}{*}{\textbf{VFM}} & \multicolumn{8}{c}{\textbf{T1n$\rightarrow$T2w}} & \multicolumn{8}{c}{\textbf{T2w$\rightarrow$T1n}} \\
\cmidrule(lr){3-10} \cmidrule(lr){11-18}
& & \multicolumn{4}{c}{\textbf{DICE (\%, mean $\pm$ std)}} & \multicolumn{4}{c}{\textbf{ASD (mm, mean $\pm$ std)}} & \multicolumn{4}{c}{\textbf{DICE (\%, mean $\pm$ std)}} & \multicolumn{4}{c}{\textbf{ASD (mm, mean $\pm$ std)}} \\
\cmidrule(lr){3-6} \cmidrule(lr){7-10} \cmidrule(lr){11-14} \cmidrule(lr){15-18}
& & \textbf{TC} & \textbf{SNFH} & \textbf{ET} & \textbf{RC} & \textbf{TC} & \textbf{SNFH} & \textbf{ET} & \textbf{RC} & \textbf{TC} & \textbf{SNFH} & \textbf{ET} & \textbf{RC} & \textbf{TC} & \textbf{SNFH} & \textbf{ET} & \textbf{RC} \\ 
\midrule

Baseline & $\times$ & 19.7$\pm$29.1 & 7.5$\pm$7.4 & 4.7$\pm$5.6 & 30.0$\pm$24.9 & 40.6$\pm$17.8 &21.2$\pm$8.6 & 40.2$\pm$18.8 & 58.9$\pm$25.7 & 4.6$\pm$4.8 & 13.9$\pm$10.5 & 3.2$\pm$4.9 & 25.9$\pm$38.1 & 48.4$\pm$22.9 & 19.4$\pm$7.6 & 38.1$\pm$15.9 & 57.1$\pm$20.6 \\

Supervised & $\times$ & 61.9$\pm$20.8 & 84.3$\pm$7.3 & 68.0$\pm$14.8 & 86.6$\pm$10.9 & 2.9$\pm$5.1 & 1.3$\pm$0.5 & 1.8$\pm$1.9 & 1.0$\pm$3.0 &62.8$\pm$20.7 &83.6$\pm$6.7& 71.6$\pm$14.9& 83.9$\pm$14.3&2.3$\pm$2.5 & 1.4$\pm$0.6 & 1.8$\pm$2.4 &1.2$\pm$2.9 \\  \midrule

UPL\cite{upl} & $\times$ & \textcolor{red}{\textbf{47.2$\pm$23.4}} & \textcolor{blue}{\textbf{36.9$\pm$22.5}} & \textcolor{red}{\textbf{61.7$\pm$48.6}} & 46.7$\pm$23.5 & 33.6$\pm$22.8 & \textcolor{blue}{\textbf{33.5$\pm$20.8}} & \textcolor{blue}{\textbf{6.8$\pm$6.1}} & \textcolor{blue}{\textbf{29.5$\pm$23.1}} & \textcolor{red}{\textbf{46.5$\pm$34.1}} & 10.7$\pm$12.1 & \textcolor{red}{\textbf{60.3$\pm$48.9}} & \textcolor{blue}{\textbf{38.7$\pm$24.6}} & \textcolor{blue}{\textbf{16.5$\pm$12.8}} & {56.9$\pm$22.1} & \textcolor{blue}{\textbf{23.2$\pm$14.3}} & \textcolor{blue}{\textbf{22.9$\pm$17.0}}\\

ProtoContra\cite{protocontra} & $\times$ & 0.7$\pm$2.3 & 4.9$\pm$12.3 & 6.1$\pm$1.9 & 15.9$\pm$17.8 & \textcolor{blue}{\textbf{19.7$\pm$14.3}} & 52.2$\pm$16.4 & 30.6$\pm$5.6 & {35.0$\pm$16.8} & 0.7$\pm$1.6 & 11.6$\pm$15.1 & 0.4$\pm$3.6 & 0.7$\pm$0.6 & {55.8$\pm$17.5} & 61.4$\pm$22.7 & 36.2$\pm$21.7 & 52.7$\pm$31.6 \\

IAPC\cite{IAPC} & $\times$ & 0.4$\pm$2.3 & {7.2$\pm$10.3} & 2.6$\pm$7.5 & 2.6$\pm$8.2 & 75.7$\pm$42.8 & {72.2$\pm$50.4} & 71.6$\pm$48.3 & 68.7$\pm$44.3 & 0.7$\pm$3.4 & {7.1$\pm$11.1} & 2.2$\pm$6.8 & 3.5$\pm$10.3 & 71.9$\pm$35.7 & 74.3$\pm$52.9 & 66.7$\pm$40.8 & 60.5$\pm$35.2 \\

DFG\cite{DFG} & \checkmark & {1.3$\pm$1.9} & 3.6$\pm$8.2 & 5.1$\pm$3.8 & 1.6$\pm$7.7 & {50.7$\pm$29.6} & 67.3$\pm$20.5 & {60.7$\pm$23.5} & 53.2$\pm$21.1 & {0.8$\pm$0.6} & 7.6$\pm$14.3 & {4.2$\pm$2.3} & {1.6$\pm$1.2} & 53.9$\pm$22.8 & 63.2$\pm$23.3 & {59.4$\pm$28.2} & {65.7$\pm$22.7} \\

IPLC\cite{iplc} & \checkmark & 17.8$\pm$3.6 & 20.5$\pm$5.2 & 29.3$\pm$11.2 & \textcolor{blue}{\textbf{52.8$\pm$9.1}} & 27.0$\pm$14.8 & 48.1$\pm$36.3 & 38.1$\pm$16.4 & 53.0$\pm$19.8 & 25.7$\pm$16.8 & \textcolor{blue}{\textbf{30.0$\pm$12.3}} & 39.7$\pm$32.9 & 16.2$\pm$17.4 & 47.2$\pm$46.1 & 36.7$\pm$23.2 & 36.3$\pm$17.6 & 45.4$\pm$17.7 \\ 

SRPL\cite{SRPL} & \checkmark  & 0.0$\pm$0.0 & 13.3$\pm$13.3 & 10.7$\pm$12.2 & 3.3$\pm$3.3 & 63.2$\pm$33.2 & 35.2$\pm$24.1 & 73.3$\pm$34.5 & 55.4$\pm$22.1 & 8.4$\pm$13.5 & 9.2$\pm$7.7 & 9.9$\pm$10.7 & 2.5$\pm$2.3 & 70.8$\pm$30.0 & \textcolor{blue}{\textbf{25.9$\pm$9.8}} & 50.7$\pm$24.4 & 51.3$\pm$13.8\\ \midrule

\textbf{Ours} & \checkmark & \textcolor{blue}{\textbf{41.6$\pm$21.0}} & \textcolor{red}{\textbf{79.9$\pm$10.1}} & \textcolor{blue}{\textbf{45.4$\pm$25.4}} & \textcolor{red}{\textbf{79.5$\pm$18.3}} & \textcolor{red}{\textbf{5.9$\pm$9.6}} & \textcolor{red}{\textbf{1.7$\pm$1.2}} &\textcolor{red}{\textbf{4.5$\pm$5.2}} & \textcolor{red}{\textbf{2.1$\pm$5.0}} & \textcolor{blue}{\textbf{41.7$\pm$26.7}} & \textcolor{red}{\textbf{78.2$\pm$11.5}} & \textcolor{blue}{\textbf{51.5$\pm$23.2}} & \textcolor{red}{\textbf{77.3$\pm$21.9}} & \textcolor{red}{\textbf{5.2$\pm$6.8}} & \textcolor{red}{\textbf{1.7$\pm$1.0}} & \textcolor{red}{\textbf{4.0$\pm$4.8}} & \textcolor{red}{\textbf{2.2$\pm$4.6}} \\

\midrule
\midrule

\multirow{3}{*}{\textbf{Methods}} & \multirow{3}{*}{\textbf{VFM}} & \multicolumn{8}{c}{\textbf{T1c$\rightarrow$T2f}} & \multicolumn{8}{c}{\textbf{T2f$\rightarrow$T1c}} \\
\cmidrule(lr){3-10} \cmidrule(lr){11-18}
& & \multicolumn{4}{c}{\textbf{DICE (\%, mean $\pm$ std)}} & \multicolumn{4}{c}{\textbf{ASD (mm, mean $\pm$ std)}} & \multicolumn{4}{c}{\textbf{DICE (\%, mean $\pm$ std)}} & \multicolumn{4}{c}{\textbf{ASD (mm, mean $\pm$ std)}} \\
\cmidrule(lr){3-6} \cmidrule(lr){7-10} \cmidrule(lr){11-14} \cmidrule(lr){15-18}
& & \textbf{TC} & \textbf{SNFH} & \textbf{ET} & \textbf{RC} & \textbf{TC} & \textbf{SNFH} & \textbf{ET} & \textbf{RC} & \textbf{TC} & \textbf{SNFH} & \textbf{ET} & \textbf{RC} & \textbf{TC} & \textbf{SNFH} & \textbf{ET} & \textbf{RC} \\ 
\midrule

Baseline & $\times$ & 4.8$\pm$5.7 & 1.3$\pm$1.4 & 14.2$\pm$15.1 & 49.3$\pm$33.5 & 39.9$\pm$30.7 & 22.7$\pm$7.9 &33.8$\pm$19.2 & 26.3$\pm$26.9 & 18.8$\pm$19.9 & 2.1$\pm$3.6 & 42.7$\pm$25.7 & 54.0$\pm$33.1 & 10.8$\pm$13.9 & 23.8$\pm$18.4 & 6.4$\pm$7.7 & 11.6$\pm$19.6 \\

Supervised & $\times$ & 63.1$\pm$20.2 & 90.3$\pm$5.6 &66.6$\pm$15.2 &83.7$\pm$11.4&2.5$\pm$2.5& 0.97$\pm$0.4& 2.2$\pm$2.7 &1.0$\pm$0.7& 82.4$\pm$15.3 & 83.6$\pm$6.9& 84.8$\pm$12.5&85.3$\pm$14.1 & 1.4$\pm$5.0& 1.4$\pm$0.5&0.8$\pm$0.9 &1.0$\pm$1.5 \\ \midrule

UPL\cite{upl} & $\times$ &\textcolor{blue}{\textbf{25.0$\pm$22.0}} &46.6$\pm$19.4 &22.6$\pm$16.3 &\textcolor{blue}{\textbf{55.9$\pm$30.2}}& \textcolor{blue}{\textbf{32.9$\pm$22.5}} & {26.1$\pm$23.8} &36.9$\pm$24.0 &29.7$\pm$26.6 & \textcolor{blue}{\textbf{43.4$\pm$27.1}} & 15.6$\pm$15.6 & \textcolor{blue}{\textbf{44.6$\pm$26.1}} & \textcolor{blue}{\textbf{53.7$\pm$29.1}} & \textcolor{blue}{\textbf{28.6$\pm$25.5}} & 46.2$\pm$22.2 & \textcolor{blue}{\textbf{27.6$\pm$23.9}} & \textcolor{blue}{\textbf{29.7$\pm$26.8}} \\

ProtoContra\cite{protocontra} & $\times$ & {0.1$\pm$1.3} & \textcolor{blue}{\textbf{58.4$\pm$27.5}} & 12.1$\pm$5.6 & 14.4$\pm$25.5 & {51.5$\pm$27.6}& \textcolor{blue}{\textbf{18.0$\pm$19.8}} & 38.0$\pm$21.4 & \textcolor{blue}{\textbf{25.8$\pm$20.5}} & 2.6$\pm$9.8 & 15.5$\pm$17.9 & 11.6$\pm$21.1 & 13.0$\pm$26.0 &{53.1$\pm$24.4} & 50.0$\pm$24.8 & 47.8$\pm$25.4 & 34.1$\pm$27.4 \\

IAPC\cite{IAPC} & $\times$ & 1.0$\pm$4.6 & 6.9$\pm$10.5 & 2.3$\pm$7.5 & 29.4$\pm$9.2 & 69.1$\pm$35.7 & 68.8$\pm$48.2 & 59.7$\pm$35.4 & 62.1$\pm$37.6 & 13.2$\pm$5.0 & 6.7$\pm$11.1 & 4.2$\pm$8.9 & 3.8$\pm$10.0 & 65.6$\pm$32.6 & 77.5$\pm$58.3 & 64.9$\pm$43.3 & 80.6$\pm$60.1 \\

DFG\cite{DFG} & \checkmark & {0.2$\pm$3.2} & 9.2$\pm$12.5 & {6.6$\pm$3.1} & {4.3$\pm$12.9} & {65.0$\pm$20.2} & 49.1$\pm$22.0 & {58.6$\pm$24.5} & {50.7$\pm$27.1} & {1.3$\pm$1.8} & {3.6$\pm$8.2} & {5.1$\pm$3.1} & {1.6$\pm$7.7} & {50.7$\pm$29.6} & 67.3$\pm$20.5 & {60.7$\pm$23.5} & 53.2$\pm$21.0 \\

IPLC\cite{iplc} & \checkmark & 17.6$\pm$8.3 & 42.8$\pm$15.6 & \textcolor{blue}{\textbf{39.9$\pm$6.1}} & 28.7$\pm$13.2 & 55.1$\pm$33.4 & 48.2$\pm$42.9 & 34.6$\pm$33.9 & 27.2$\pm$16.1 & 29.6$\pm$24.4 & \textcolor{blue}{\textbf{39.9$\pm$18.1}} & 13.7$\pm$7.4 & 44.8$\pm$18.6 & 43.9$\pm$28.1 & \textcolor{blue}{\textbf{19.2$\pm$8.4}} & 59.4$\pm$47.0 & 33.0$\pm$19.9 \\

SRPL\cite{SRPL} & \checkmark & 15.9$\pm$18.2 & 3.8$\pm$4.1 & 23.3$\pm$18.6 & 5.5$\pm$8.2 & 40.2$\pm$35.1 & 30.6$\pm$12.5 & \textcolor{blue}{\textbf{34.5$\pm$30.9}} & 30.2$\pm$24.3 & 24.1$\pm$20.7 & 7.4$\pm$7.0 & 22.3$\pm$20.2 & 6.5$\pm$8.0 & 36.4$\pm$46.0 & 27.7$\pm$12.3 & 45.0$\pm$37.5 & 39.0$\pm$21.5 \\ \midrule

\textbf{Ours} & \checkmark & \textcolor{red}{\textbf{42.2$\pm$26.1}} & \textcolor{red}{\textbf{86.0$\pm$10.1}} & \textcolor{red}{\textbf{44.6$\pm$24.5}} & \textcolor{red}{\textbf{77.1$\pm$19.7}} & \textcolor{red}{\textbf{7.4$\pm$9.0}} & \textcolor{red}{\textbf{1.2$\pm$0.6}} & \textcolor{red}{\textbf{4.5$\pm$4.7}} & \textcolor{red}{\textbf{1.9$\pm$3.0}} & \textcolor{red}{\textbf{72.8$\pm$23.7}} & \textcolor{red}{\textbf{78.6$\pm$10.4}} & \textcolor{red}{\textbf{78.3$\pm$19.8}} & \textcolor{red}{\textbf{78.2$\pm$20.1}} & \textcolor{red}{\textbf{2.5$\pm$3.4}} & \textcolor{red}{\textbf{1.7$\pm$0.7}} & \textcolor{red}{\textbf{1.2$\pm$1.5}} & \textcolor{red}{\textbf{1.8$\pm$3.4}} \\ 

\bottomrule
\end{tabular}%
}
\vspace{-3mm}
\end{table*}
We validate our method on AMOS \cite{amos}, BraTS \cite{brats2024}, CAMUS \cite{CAMUS}, ACDC \cite{acdc}, Kvasir \cite{kvasir}, and CVCDB \cite{CVCDB} datasets. \textbf{AMOS} consists of 500 CT and 100 MRI scans collected from a multi-center, multi-vendor, multi-modality, multi-phase, and multi-disease cohort, with voxel-level annotations for 15 abdominal organs. \textbf{BraTS} provides 2200 3D MRI scans across four modalities: T1n, T1c, T2w, and T2f, which include segmentation targets for four subregions: enhancing tissue (ET), tumor core (TC), surrounding non-enhancing FLAIR hyperintensity (SNFH), and resection cavity (RC). \textbf{CAMUS} provides 2D apical four-chamber and two-chamber view sequences acquired from 500 patients, with pixel-level annotations for the left ventricle (LV), myocardium (MYO), and left atrium. \textbf{ACDC} contains 150 MRI scans categorized into five subgroups, which include slices covering the LV, MYO, and right ventricle from base to apex. \textbf{Kvasir} is a widely used benchmark dataset consisting of 1,000 endoscopic images with pixel-wise annotated polyp segmentation. \textbf{CVCDB} comprises 612 frames extracted from multiple colonoscopy video sequences with corresponding masks delineating the polyp regions. Notably, in US$\rightarrow$MR and MR$\rightarrow$US, we restrict our evaluation to LV and MYO from ACDC and CAMUS to maintain consistency in the targets. 

We evaluate Tell2Adapt on 10 domain adaptation directions: T1n$\rightarrow$T2w, T2w$\rightarrow$T1n, T1c$\rightarrow$T2f and T2f$\rightarrow$T1c for brain targets on BraTS, MR$\rightarrow$CT and CT$\rightarrow$MR for abdominal targets on AMOS, Kvasir$\rightarrow$CVCDB and CVCDB$\rightarrow$Kvasir for polyp on CVCDB and Kvasir, and US$\rightarrow$MR and MR$\rightarrow$US for cardiac targets on ACDC and CAMUS. The dataset is randomly slice-wise divided into training and testing sets in an 8:2 ratio, and the performance is quantitatively evaluated using DICE and ASD.

All experiments were implemented using PyTorch and conducted on two NVIDIA RTX 4090 GPUs. For the source model, we used the nnUNet framework\cite{nnunet}, which employs a ResNet-based U-Net architecture, trained for 500 epochs. For all the comparison methods, we follow the official implementation and hyperparameters provided in their original paper. In CAPR, we employ Qwen3-VL-8B-Instruct~\cite{qwen3}, using a unified meta-prompt that remains identical across all adaptation directions, as detailed in the Appendix. For VPR, the anatomical priors are sourced from the official HuggingFace repository of BiomedParse and are also included in the supplementary materials, where we additionally provide all implementation details and the complete code of Tell2Adapt to ensure reproducibility.
\subsection{Comparison with SOTA Methods}
To further assess the effectiveness of our framework, we compare Tell2Adapt with six SOTA SFUDA methods, covering both general-purpose approaches (IAPC) and medical-purpose approaches (UPL, ProtoContra, IPLC, SRPL, and DFG). Notably, our evaluation aligns with the emerging paradigm of VFM-guided SFUDA, where DFG and IPLC utilize MedSAM, while SRPL is built upon SAM. The remaining methods represent traditional SFUDA that do not utilize VFM. In our experiments, Baseline denotes the source model directly tested on the target domain without adaptation, while Supervised refers to training and testing on the target domain with full supervision.

\textbf{Results on Abdominal Targets}
As detailed in \Cref{tab:AMOS}, Tell2Adapt demonstrates exceptional performance in abdominal targets. In MR$\rightarrow$CT, our method achieves a mean DICE of 88.2\%. Remarkably, this result is not only a substantial improvement over Baseline but is also on par with the Supervised (88.4\%), indicating a high-performing adaptation without any target labels.

The framework's robustness is further confirmed in CT$\rightarrow$MR. Here, Tell2Adapt attains a highly competitive mean DICE of 82.3\%, once again closely approaching the Supervised upper bound of 86.2\%. This consistent, high-level performance across both demanding adaptation on a complex multi-organ benchmark validates the powerful generalization capability of Tell2Adapt, effectively bridging the domain shift between MR and CT.

\textbf{Results on Brain Targets}
\Cref{tab:BraTS} presents the comprehensive performance of Tell2Adapt on BraTS, evaluated across four domain adaptation directions. In T1n$\rightarrow$T2w, our method achieves a leading DICE of 41.6\% on TC, 79.9\% on SNFH, 45.4\% on ET, and 79.5\% on RC. This represents a significant improvement over Baseline. The strong performance, particularly on SNFH and RC, highlights our method's ability to handle large structural targets robustly. This strong performance is maintained in T2w$\rightarrow$T1n, where Tell2Adapt again shows a strong overall profile with DICE of 78.2\% on SNFH, 51.5\% on ET, and 77.3\% on RC. While UPL achieves a high DICE on TC and ET, our method provides a more balanced performance across DICE and ASD.

The advantage of our framework is even more pronounced in the post-contrast adaptation. In T1c$\rightarrow$T2f, Tell2Adapt achieves a DICE of 86.0\% and an ASD of 1.2 mm on SNFH, surpassing all competing methods by a substantial margin. Similarly, in T2f$\rightarrow$T1c, our method demonstrates remarkable consistency with high DICE across all subregions: 72.8\% on TC, 78.6\% on SNFH, 78.3\% on ET, and 78.2\% on RC. Critically, these gains in overlap are complemented by superior boundary precision, as our method also achieves the lowest ASD on all targets in these adaptation directions.

\textbf{Results on Cardiac Targets}
We also evaluate Tell2Adapt on the extremely challenging cross-modality cardiac adaptation: MR$\rightarrow$US and US$\rightarrow$MR. These tasks represent a profound domain shift due to the fundamentally different imaging physics of MR and US. As shown in \Cref{tab:Cardic}, the severe domain gap causes the Baseline to fail, resulting in near-zero DICE and exceedingly large ASD. Existing SFUDA methods also fail to bridge this gap, producing almost meaningless segmentation. In contrast, Tell2Adapt exhibits strong robustness. For MR$\rightarrow$US, our framework achieves DICE of 94.6\% on LV and 88.5\% on MYO, closely approaching Supervised. This demonstrates that incorporating VFM knowledge can effectively mitigate extreme domain shifts. Similar trends are observed in US$\rightarrow$MR, where Tell2Adapt again attains near Supervised performance with DICE of 95.5\% on LV and 89.2\% on MYO. The consistently lowest ASD further confirms the high accuracy and anatomical precision of the predicted segmentation boundaries.
\begin{table}[t]
\centering
\caption{DICE and ASD of segmentation results on cardiac targets.}
\vspace{-3mm}
\label{tab:Cardic}
\resizebox{\columnwidth}{!}{%
\renewcommand{\arraystretch}{0.75}
\begin{tabular}{@{}c@{\hspace{0.3cm}}c cccc cccc@{}}
\toprule
\multirow{3}{*}{\textbf{Methods}} & \multirow{3}{*}{\textbf{VFM}} & \multicolumn{4}{c}{\textbf{MR $\rightarrow$ US}} & \multicolumn{4}{c}{\textbf{US $\rightarrow$ MR}} \\
\cmidrule(lr){3-6} \cmidrule(lr){7-10}
& & \multicolumn{2}{c}{\textbf{DICE (\%, mean $\pm$ std)}} & \multicolumn{2}{c}{\textbf{ASD (mm, mean $\pm$ std)}} & \multicolumn{2}{c}{\textbf{DICE (\%, mean $\pm$ std)}} & \multicolumn{2}{c}{\textbf{ASD (mm, mean $\pm$ std)}} \\
\cmidrule(lr){3-4} \cmidrule(lr){5-6} \cmidrule(lr){7-8} \cmidrule(lr){9-10}
& & \textbf{LV} & \textbf{MYO} & \textbf{LV} & \textbf{MYO} & \textbf{LV} & \textbf{MYO} & \textbf{LV} & \textbf{MYO} \\ 
\midrule
Baseline & $\times$  & 1.7$\pm$0.7 & 3.1$\pm$0.3 & 56.2$\pm$37.2 & 66.9$\pm$42.3& 3.2$\pm$0.6 & 1.8$\pm$0.3 & 60.5$\pm$34.9 & 56.8$\pm$27.5 \\
Supervised & $\times$ & 94.4$\pm$2.7 & 88.1$\pm$4.6 & 3.2$\pm$1.6 & 4.1$\pm$1.3 & 95.8$\pm$3.8 & 91.2$\pm$2.1 & 0.3$\pm$0.1 & 0.3$\pm$0.1 \\ \midrule
UPL \citep{upl} & $\times$ & 2.8$\pm$0.1 & 0.0$\pm$0.0 & 90.5$\pm$3.2 & 0.0$\pm$0.0 & 0.5$\pm$0.4 & 2.9$\pm$2.1 & 88.9$\pm$9.6 & 85.6$\pm$11.6 \\
ProtoContra \citep{protocontra} & $\times$ & 1.2$\pm$1.2 & \textcolor{blue}{\textbf{17.0$\pm$7.2}} & 54.9$\pm$14.2 & 56.4$\pm$13.9 & 5.2$\pm$5.8 & 4.9$\pm$3.2 & 72.9$\pm$12.4 & 74.8$\pm$10.5 \\
IAPC \citep{IAPC} & $\times$ & \textcolor{blue}{\textbf{18.4$\pm$14.5}} & 3.1$\pm$7.5 & 51.3$\pm$29.6 & 97.9$\pm$45.3 & \textcolor{blue}{\textbf{25.8$\pm$15.2}} & 5.6$\pm$0.8 & \textcolor{blue}{\textbf{34.1$\pm$13.9}} & 42.6$\pm$12.7 \\
DFG \citep{DFG} & \checkmark & 4.6$\pm$4.1 & 4.5$\pm$3.7 & 51.6$\pm$12.9 & 63.1$\pm$14.3 & 2.3$\pm$1.7 & 6.9$\pm$4.4 & 69.3$\pm$16.7 & 66.5$\pm$18.3 \\
IPLC \citep{iplc} & \checkmark & 1.7$\pm$2.0 & 5.8$\pm$2.5 & \textcolor{blue}{\textbf{50.5$\pm$18.1}} & \textcolor{blue}{\textbf{48.3$\pm$18.9}} & 2.9$\pm$3.6 & \textcolor{blue}{\textbf{11.5$\pm$8.9}} & 48.4$\pm$11.2 & \textcolor{blue}{\textbf{26.1$\pm$10.5}} \\

SRPL \citep{SRPL} & \checkmark & 0.7$\pm$1.1 & 0.9$\pm$1.2 & 83.1$\pm$31.0 & 79.8$\pm$19.2 & 8.5$\pm$9.3 & 9.1$\pm$9.3 & 30.1$\pm$15.3 & 45.9$\pm$22.0 \\ \midrule

\textbf{Ours} & \checkmark & \textcolor{red}{\textbf{94.6$\pm$2.6}} & \textcolor{red}{\textbf{88.5$\pm$4.5}} & \textcolor{red}{\textbf{3.2$\pm$1.5}} & \textcolor{red}{\textbf{4.1$\pm$1.6}} & \textcolor{red}{\textbf{95.5$\pm$3.7}} & \textcolor{red}{\textbf{89.2$\pm$3.9}} & \textcolor{red}{\textbf{3.1$\pm$0.2}} & \textcolor{red}{\textbf{4.1$\pm$0.4}}  \\ 
\bottomrule
\end{tabular}%
}
\vspace{-3mm}
\end{table}
\begin{table}[t]
\centering
\caption{DICE and ASD of segmentation results on polyp.}
\label{tab:Polyp}
\vspace{-3mm}
\resizebox{\columnwidth}{!}{%
\renewcommand{\arraystretch}{0.75}
\small
\begin{tabular}{@{}c@{\hspace{0.3cm}}c cc cc@{}}
\toprule
\multirow{3}{*}{\textbf{Methods}} & \multirow{3}{*}{\textbf{VFM}} & \multicolumn{2}{c}{\textbf{Kvasir $\rightarrow$ CVCDB}} & \multicolumn{2}{c}{\textbf{CVCDB $\rightarrow$ Kvasir}} \\
\cmidrule(lr){3-4} \cmidrule(lr){5-6}
& & \textbf{DICE (\%, mean $\pm$ std)} & \textbf{ASD (mm, mean $\pm$ std)} & \textbf{DICE (\%, mean $\pm$ std)} & \textbf{ASD (mm, mean $\pm$ std)} \\
\midrule
Baseline & $\times$ & 65.1$\pm$21.5 & 5.8$\pm$11.1 & 60.2$\pm$27.0& 23.7$\pm$20.4 \\

Supervised & $\times$ & 99.8$\pm$0.1 & 0.36$\pm$0.1 & 97.4$\pm$1.3 & 0.9$\pm$1.1 \\ \midrule

UPL \citep{upl} & $\times$ & 67.1$\pm$14.6 &  \textcolor{blue}{\textbf{2.2$\pm$0.9}} & 60.2$\pm$31.3 & 11.4$\pm$17.2 \\

ProtoContra \citep{protocontra} & $\times$ & \textcolor{blue}{\textbf{78.0$\pm$12.1}} & \textcolor{red}{\textbf{0.9$\pm$0.6}} & \textcolor{blue}{\textbf{68.9$\pm$26.4}} & \textcolor{blue}{\textbf{9.5$\pm$15.0}} \\

IAPC \citep{IAPC} & $\times$ & 73.3$\pm$29.6 & 4.4$\pm$7.9 & 61.2$\pm$26.3 & 13.7$\pm$13.2 \\

DFG \citep{DFG} & \checkmark & 19.1$\pm$14.7 & 97.5$\pm$17.6 & 26.8$\pm$18.1 & 87.5$\pm$19.5 \\

IPLC \citep{iplc} & \checkmark & 34.8$\pm$11.7 & 34.7$\pm$22.9 & 45.3$\pm$21.2 & 19.2$\pm$8.4 \\

SRPL \citep{SRPL} & \checkmark & 33.0$\pm$22.7 & 31.4$\pm$19.6 & 44.9$\pm$26.0 & 18.9$\pm$16.9 \\ \midrule

\textbf{Ours} & \checkmark & \textcolor{red}{\textbf{89.1$\pm$16.0}} & {4.7$\pm$6.6} & \textcolor{red}{\textbf{88.3$\pm$14.7}} & \textcolor{red}{\textbf{6.2$\pm$3.2}}  \\ 
\bottomrule
\end{tabular}%
}
\vspace{-3mm}
\end{table}
\begin{table*}[!t]
\centering
\caption{DICE (\%, mean $\pm$ std) of segmentation results in the ablation study on abdominal targets. The corresponding ASD results for all anatomical targets are provided in the Appendix, along with detailed descriptions of the chaotic prompts and normal prompts.}
\label{tab:Ablation}
\resizebox{2\columnwidth}{!}{%
\begin{tabular}{@{}cccccccccccccccccc@{}}
\toprule
\multicolumn{17}{c}{\textbf{MR$\rightarrow$CT}} \\ \midrule
{ \textbf{Methods}} & { \textbf{Spleen}} & { \textbf{R. K}} & { \textbf{L. K}} & { \textbf{Gallb}} & { \textbf{Esoph}} & { \textbf{Liver}} & { \textbf{Stom}} & { \textbf{Aortic}} & { \textbf{V-Cava}} & { \textbf{Pancr}} & { \textbf{R. Adre}} & { \textbf{L. Adre}} & { \textbf{Duod}} & { \textbf{Bladder}} & { \textbf{Pros/Uter}} & { \textbf{mDICE}}\\ \midrule
{ Baseline} & { 65.1$\pm$30.4} & { 71.2$\pm$26.9} & { 63.2$\pm$29.6} & { 27.2$\pm$27.7} & {42.2$\pm$22.1} & {82.2$\pm$9.9} & {46.9$\pm$28.6} & {72.0$\pm$20.3} & {61.7$\pm$23.7} & {55.7$\pm$23.3} & {39.7$\pm$22.8} & {44.7$\pm$24.2} & {38.5$\pm$22.4} & {0.0$\pm$0.0} & {0.0$\pm$0.0} & {47.4$\pm$24.7}\\ \midrule
Ours w/o CAPR (Chaotic Prompts)& 85.2$\pm$1.5 &0.0$\pm$0.0 &88.3$\pm$2.5& 84.4$\pm$10.0 &{84.7$\pm$9.4} &0.0$\pm$0.0& 0.0$\pm$0.0& 85.4$\pm$2.1& 69.7$\pm$13.7& 0.0$\pm$0.0& 70.5$\pm$7.6&0.0$\pm$0.0&\textcolor{red}{\textbf{81.7$\pm$6.1}}& 0.0$\pm$0.0 &{83.7$\pm$8.2} &48.9$\pm$37.1\\
{ Ours w/o CAPR (Normal Prompts)} & {91.1$\pm$2.2} & {92.0$\pm$4.7} & {89.9$\pm$3.3} & {84.6$\pm$7.2} & \textcolor{red}{\textbf{87.5$\pm$2.1}} & {92.4$\pm$3.5} & {85.8$\pm$4.9} & {88.4$\pm$5.2} & {90.2$\pm$5.7} & {81.1$\pm$6.8} & {75.4$\pm$9.8} & {73.9$\pm$3.7} & {78.2$\pm$5.4} & \textcolor{blue}{\textbf{83.4$\pm$6.0}} & \textcolor{red}{\textbf{88.3$\pm$7.2}} & {85.5$\pm$6.7}\\
{ Ours w/o VPR (Normal Prompts)} & { 90.1$\pm$2.3} & { 91.8$\pm$3.7} & {92.5$\pm$4.4} & {80.2$\pm$12.1} & { 83.2$\pm$6.1} & { 94.9$\pm$3.3} & {90.0$\pm$7.6} & {91.8$\pm$4.1} & { 90.1$\pm$5.8} & { 80.6$\pm$11.6} & { 72.7$\pm$8.0} & { 74.6$\pm$9.7} & { 75.4$\pm$16.8} & { 82.3$\pm$14.4} & \textcolor{blue}{\textbf{ 84.9$\pm$7.0}} & { 85.0$\pm$7.4}\\ \midrule
{ Ours  (Chaotic Prompts) } & {\textcolor{blue}{\textbf{96.9$\pm$1.4}}} & {\textcolor{blue}{\textbf{96.0$\pm$1.3}}} & {\textcolor{blue}{\textbf{95.7$\pm$1.5}}} & {\textcolor{red}{\textbf{85.4$\pm$10.2}}} & {\textcolor{blue}{\textbf{85.2$\pm$6.6}}} & {\textcolor{blue}{\textbf{96.8$\pm$7.8}}} & {\textcolor{blue}{\textbf{90.5$\pm$10.8}}} & {\textcolor{blue}{\textbf{93.3$\pm$2.1}}} & {\textcolor{red}{\textbf{91.5$\pm$4.0}}} & {\textcolor{red}{\textbf{85.6$\pm$5.6}}} & {\textcolor{blue}{\textbf{77.3$\pm$9.9}}} & {\textcolor{red}{\textbf{78.4$\pm$11.3}}} & {\textcolor{blue}{\textbf{80.8$\pm$6.1}}} & {83.1$\pm$11.0} & {49.2$\pm$7.6} & {\textcolor{blue}{\textbf{85.7$\pm$12.6}}}\\ 
{ Ours (Normal Prompts)} & {\textcolor{red}{\textbf{97.1$\pm$1.6}}} & {\textcolor{red}{\textbf{96.3$\pm$2.9}}} & {\textcolor{red}{\textbf{96.2$\pm$4.6}}} & {\textcolor{blue}{\textbf{85.2$\pm$16.4}}} & { 84.0$\pm$8.6} & {\textcolor{red}{\textbf{96.9$\pm$2.6}}} & {\textcolor{red}{\textbf{91.9$\pm$7.5}}} & {\textcolor{red}{\textbf{94.9$\pm$2.2}}} & {\textcolor{blue}{\textbf{90.5$\pm$5.5}}} & \textcolor{blue}{\textbf{84.7$\pm$10.5}} & {\textcolor{red}{\textbf{77.4$\pm$8.1}}} & {\textcolor{blue}{\textbf{77.3$\pm$10.3}}} & { 79.7$\pm$14.2} & {\textcolor{red}{\textbf{85.8$\pm$16.4}}} & {84.6$\pm$9.2} & {\textcolor{red}{\textbf{88.2$\pm$7.1}}}\\ 
\bottomrule
\end{tabular}%
}

\resizebox{2\columnwidth}{!}{%
\begin{tabular}{@{}cccccccccccccccc@{}}
\toprule
\multicolumn{15}{c}{\textbf{CT$\rightarrow$MR}} \\ \midrule
{ \textbf{Methods}} & { \textbf{Spleen}} & { \textbf{R. K}} & { \textbf{L. K}} & { \textbf{Gallb}} & { \textbf{Esoph}} & { \textbf{Liver}} & { \textbf{Stom}} & { \textbf{Aortic}} & { \textbf{V-Cava}} & { \textbf{Pancr}} & { \textbf{R. Adre}} & { \textbf{L. Adre}} & { \textbf{Duod}} & { \textbf{mDICE}}\\ \midrule
{ Baseline} & { 58.6$\pm$47.0} & { 86.6$\pm$9.2} & { 89.6$\pm$0.7} & {53.1$\pm$28.8} & 58.9$\pm$34.2 & {73.4$\pm$33.0} & 60.0$\pm$19.5 & 61.4$\pm$29.8 & 64.0$\pm$25.1 &65.0$\pm$19.2 & 56.5$\pm$3.3 & 64.7$\pm$10.2 & 44.3$\pm$25.0&64.3$\pm$12.5\\ \midrule
{ Ours w/o CAPR (Chaotic Prompts)} & {85.7$\pm$1.4} & {0.0$\pm$0.0} & {90.6$\pm$2.6} & {69.5$\pm$5.7} & {70.6$\pm$7.7} & {0.0$\pm$0.0} & {0.0$\pm$0.0} & {90.9$\pm$1.8} & {42.0$\pm$18.9} & {0.0$\pm$0.0} & {54.3$\pm$9.4} & {0.0$\pm$0.0} & {\textcolor{red}{\textbf{70.7$\pm$5.1}}} & {44.2$\pm$36.9}\\
{ Ours w/o CAPR (Normal Prompts)} & {92.5$\pm$4.4} & {91.8$\pm$7.8} & {90.4$\pm$2.9} & {72.6$\pm$8.5} & \textcolor{red}{\textbf{74.4$\pm$7.1}} & {89.1$\pm$4.6} & {86.2$\pm$9.7} & {91.4$\pm$2.2} & {80.5$\pm$5.9} & {77.4$\pm$8.1} & {61.3$\pm$4.9} & {60.2$\pm$9.8} & \textcolor{blue}{\textbf{70.1$\pm$1.2}}& {79.8$\pm$11.6}\\
{ Ours w/o VPR (Normal Prompts) } & {90.1$\pm$4.1} & {92.2$\pm$2.7} & {91.2$\pm$3.8} & {70.3$\pm$9.5} & {66.8$\pm$9.8} & {93.5$\pm$3.6} & {85.3$\pm$10.2} & {87.5$\pm$2.3} & {83.9$\pm$6.2} & \textcolor{blue}{\textbf{80.4$\pm$5.5}} & \textcolor{blue}{\textbf{62.5$\pm$5.8}} & {55.0$\pm$17.3} & {63.7$\pm$6.5} & {78.6$\pm$13.7}\\ \midrule
{ Ours (Chaotic Prompts) } & {\textcolor{blue}{\textbf{94.3$\pm$1.4}}} & {\textcolor{red}{\textbf{94.6$\pm$1.3}}} & {\textcolor{blue}{\textbf{94.1$\pm$2.6}}} & {\textcolor{red}{\textbf{76.1$\pm$7.1}}} & {70.3$\pm$7.4} & {\textcolor{red}{\textbf{97.0$\pm$0.7}}}& {\textcolor{blue}{\textbf{87.9$\pm$3.2}}} & {\textcolor{red}{\textbf{91.9$\pm$1.8}}} & {\textcolor{red}{\textbf{87.8$\pm$3.3}}} & {79.0$\pm$3.5} & {60.8$\pm$9.2} & {\textcolor{red}{\textbf{63.8$\pm$15.7}}} & {70.0$\pm$4.9}& {\textcolor{blue}{\textbf{82.1$\pm$12.4}}}\\ 
{ Ours (Normal Prompts)} & {\textcolor{red}{\textbf{94.8$\pm$3.8}}} & {\textcolor{blue}{\textbf{94.4$\pm$0.8}}} & {\textcolor{red}{\textbf{95.2$\pm$0.8}}} & {\textcolor{blue}{\textbf{73.3$\pm$7.4}}} & {\textcolor{blue}{\textbf{71.6$\pm$8.9}}} & {\textcolor{blue}{\textbf{96.8$\pm$0.4}}} & {\textcolor{red}{\textbf{88.9$\pm$7.2}}} & {\textcolor{blue}{\textbf{91.5$\pm$2.7}}} & {\textcolor{blue}{\textbf{85.7$\pm$4.3}}} & {\textcolor{red}{\textbf{85.9$\pm$5.7}}} & {\textcolor{red}{\textbf{65.5$\pm$3.0}}} & {\textcolor{blue}{\textbf{61.0$\pm$14.8}}} & {68.9$\pm$6.3}& {\textcolor{red}{\textbf{82.3$\pm$12.9}}}\\ 
\bottomrule
\end{tabular}%
}

\footnotesize{~~~~\textit{Note}:  A DICE of 0.0 indicates a complete prediction failure for the corresponding targets.}
\vspace{-3mm}
\end{table*}

\textbf{Results on Polyp Target}
We further validate Tell2Adapt on polyp segmentation, which requires adaptation between endoscopic datasets with significant variations in lighting, camera, and polyp appearance. In \Cref{tab:Polyp}, our framework demonstrates clear advantages in this challenging scenario.

In Kvasir$\rightarrow$CVCDB, our method achieves a robust DICE of 89.1\%. This represents a substantial margin of +11.1\% over the second-best method, ProtoContra (78.0\%). The performance improvement is even more pronounced in CVCDB$\rightarrow$Kvasir, where Tell2Adapt attains a DICE of 88.3\%, surpassing ProtoContra by an impressive +19.4\%. Notably, in CVCDB$\rightarrow$Kvasir, this gain in overlap accuracy is complemented by a superior boundary delineation, as indicated by a better ASD of 6.2 mm compared to 9.5 mm from the second-best method. These consistent and large performance gains across both adaptation underscore our framework's ability to distill robust, generalizable knowledge from BiomedParse, effectively overcoming domain-specific variations where other SOTA methods falter.

\textbf{Ablation Study}
\label{sec:Ablation_Study}
We conduct a comprehensive ablation study to evaluate the individual contributions of CAPR and VPR. In Table~\ref{tab:Ablation}, the importance of VPR is evident when comparing Ours (Normal Prompts) with Ours w/o VPR (Normal Prompts). Removing VPR results in a mean DICE decrease of 3.2\% (from 88.2\% to 85.0\%) in MR$\rightarrow$CT and 3.7\% (from 82.3\% to 78.6\%) in CT$\rightarrow$MR, highlighting its effectiveness in enforcing anatomical plausibility.

CAPR proves to be the most essential component, serving as both a corrective and an optimizing mechanism in the adaptation process. When the framework receives Chaotic Prompts without CAPR, performance drops drastically to 48.9\% and 44.2\% mean DICE in MR$\rightarrow$CT and CT$\rightarrow$MR, respectively, with critical organs such as the liver, stomach, and pancreas completely failing. Enabling CAPR restores the system to 85.7\% mean DICE in MR$\rightarrow$CT (+36.8\%) and 82.1\% in CT$\rightarrow$MR (+37.9\%), as demonstrated by the comparison between Ours w/o CAPR (Chaotic Prompts) and Ours (Chaotic Prompts). Beyond correction, the improvement from Ours w/o CAPR (Normal Prompts) to Ours (Normal Prompts) confirms that CAPR not only mitigates prompt noise but also refines semantic alignment, leading to further optimization. This dual effect, offering robustness through correction and optimization, together with VPR’s refinement, underscores the generality of Tell2Adapt.

\section{Discussion}
Our extensive experiments demonstrate that Tell2Adapt successfully establishes a unified and highly effective SFUDA framework, a significant step forward from prior specialized methods. This success stems from its ability to generate high-fidelity supervisory signals, overcoming the limitations of traditional SFUDA methods, which rely solely on self-training or entropy minimization from the source model, often resulting in unreliable supervision that limits their generalization and fails in high-gap scenarios.

Our approach leverages the vast knowledge of a VFM and decouples the VFM's prompt quality from the pseudo-labels of the source model. This stands in sharp contrast to other VFM-guided SFUDA methods (IPLC, DFG, and SRPL) that remain critically coupled to the source model. These methods use the source model's pseudo-labels to generate spatial prompts for VFM. This dependency creates a severe error accumulation cascade, where an inaccurate pseudo-label from the source model generates an erroneous prompt, which subsequently misguides the VFM. As a result, the predictions may degrade severely, sometimes even worse than traditional SFUDA.

Our framework leverages text prompts rather than spatial prompts. This strategy, powered by CAPR, provides a stable, semantically rich, and high-quality supervisory signal independent of the source model. It breaks the failure cascade, enabling robust adaptation even in extreme-gap scenarios. Furthermore, VPR enhances anatomical plausibility, serving as a crucial clinical reliability check. By filtering out predictions that are statistically inconsistent with general anatomical properties, it refines segmentation and reduces false positives, which are often overlooked in traditional SFUDA but are vital for real-world deployment.

\textbf{Limitations:} Despite its strong performance, Tell2Adapt has limitations. Although the adapted model is lightweight, the pseudo-label generation stage remains computationally demanding. On average, it requires 3.47\,s per abdominal volume, 1.74\,s per brain volume, 1.93\,s per cardiac volume, and 0.26\,s per polyp image. 
\section{Conclusion}
In this work, we tackled the fundamental challenge of generalization in SFUDA for medical image segmentation. We introduced Tell2Adapt, a unified framework that leverages the rich knowledge of VFM. By incorporating CAPR to ensure reliable VFM prompt and VPR to enforce anatomical plausibility, Tell2Adapt achieves SOTA across 10 adaptation directions and 22 anatomical targets. Beyond offering an effective solution, Tell2Adapt establishes a new paradigm for SFUDA.

{
    \small
    \bibliographystyle{ieeenat_fullname}
    \bibliography{ref}
}


\onecolumn
{
    \centering
    \Large
    \textbf{Tell2Adapt: A Unified Framework for Source Free Unsupervised Domain Adaptation via Vision Foundation Model}\\
    \vspace{0.5em}Appendix \\
    \vspace{1.0em}
}
\setcounter{page}{1}
\appendix
This appendix complements the main paper by providing extended quantitative evaluations and detailed implementation specifications to ensure full reproducibility. 
While the main paper focuses primarily on DICE-based comparisons in abdominal targets, this appendix further reports ASD and elaborates on the prompt regularization mechanism that supports the Tell2Adapt framework. 
The appendix is organized as follows:

\begin{itemize}
    \item \textbf{Extended Quantitative Evaluation (\Cref{sec:appendix_A}):} 
    We present the complete ASD for all abdominal organs in both the MR$\rightarrow$CT and CT$\rightarrow$MR directions, together with ASD from the ablation study. 
    These results reinforce the effectiveness of Tell2Adapt in achieving accurate and anatomically consistent segmentations.
    
    \item \textbf{Details in Ablation Study for CAPR  (\Cref{sec:appendix_B}):} 
    We describe the procedure for constructing chaos prompts used in the ablation study and provide the full list of chaos prompts. This section details how perturbations are introduced to assess the robustness of CAPR.
    
    \item \textbf{Reproducibility Details (\Cref{sec:appendix_C}):} 
    To enable exact replication, we include the meta-prompt used in CAPR for LLM and the complete set of input prompts for all anatomical targets.
\end{itemize}

\section{Extended ASD Evaluation on Abdominal Targets}
To complement DICE reported in the main paper, this section provides a comprehensive analysis of ASD performance for all abdominal targets. We report full ASD for both MR$\rightarrow$CT and CT$\rightarrow$MR directions, following the same evaluation protocol used in the main paper. 
These results offer a finer-grained assessment of boundary accuracy, which is particularly important for small and geometrically complex organs where DICE alone may not fully capture segmentation quality.

In addition to the adaptation results, we also include ASD from the ablation study to further elucidate the contribution of CAPR and VPR. Together, these quantitative evaluations provide a comprehensive assessment of Tell2Adapt's effectiveness in achieving anatomically precise segmentations under severe cross-modality domain shifts.

\label{sec:appendix_A}
\subsection{Adaptation Results in ASD for Abdominal Targets}
Due to space limitations, \Cref{tab:AMOS} in the main paper reports only DICE for abdominal segmentation. 
For completeness, we provide the corresponding ASD in \Cref{tab:AMOS_ASD_adaptation}. 
All ASD are computed over the same test subjects used in the main paper, following identical evaluation protocols and spacing normalization.
\begin{table*}[b]
\centering
\caption{ASD (mm, mean $\pm$ std) of segmentation results on abdominal targets. } 
\label{tab:AMOS_ASD_adaptation}

\resizebox{\columnwidth}{!}{%
\begin{tabular}{@{}ccccccccccccccccccc@{}}
\toprule
\multicolumn{17}{c}{\textbf{MR$\rightarrow$CT}} \\ \midrule

{ \textbf{Methods}} & { \textbf{VFM}} & { \textbf{Spleen}} & { \textbf{R. K}} & { \textbf{L. K}} & { \textbf{Gallb}} & { \textbf{Esoph}} & { \textbf{Liver}} & { \textbf{Stom}} & { \textbf{Aortic}} & { \textbf{V-Cava}} & { \textbf{Pancr}} & { \textbf{R. Adre}} & { \textbf{L. Adre}} & { \textbf{Duod}} & { \textbf{Bladder}} & { \textbf{Pros/Uter}} & { \textbf{mASD}} \\ \midrule

{ Baseline} & $\times$ & {17.5$\pm$19.2} & {17.2$\pm$18.5} & {15.5$\pm$17.8} & {23.1$\pm$30.7} & {7.8$\pm$9.6} & {11.4$\pm$7.2} & {15.3$\pm$14.6}& {8.9$\pm$9.9} & {6.9$\pm$11.2} & {9.1$\pm$10.7}  & {4.7$\pm$3.4} & {4.8$\pm$4.2} & {10.3$\pm$9.7}  & {N/A} & {N/A} & {13.0$\pm$4.8}  \\

{ Supervised} & $\times$ & {1.1$\pm$0.3} & {1.1$\pm$0.8} & {1.2$\pm$1.4} & {1.2$\pm$2.4} & {0.8$\pm$0.7} & {2.3$\pm$0.5} & {1.9$\pm$1.6}& {1.1$\pm$1.3} & {1.0$\pm$0.4} & {1.4$\pm$1.0} & {0.7$\pm$0.7} & {0.7$\pm$0.7} & {1.6$\pm$1.1}  & {1.9$\pm$1.9} & {2.7$\pm$7.7} & {1.4$\pm$0.6} \\ \midrule

{ UPL \cite{upl}} & $\times$& \textcolor{blue}{\textbf{79.0$\pm$19.9}} & {87.2$\pm$15.2} & {72.4$\pm$17.6} & {N/A}  & {14.8$\pm$17.6} & {37.0$\pm$20.6} & {31.6$\pm$21.5}& {23.7$\pm$12.8} & {50.2$\pm$15.4} & {29.3$\pm$16.7}  & {N/A} & {N/A} & {54.4$\pm$23.4}  & {N/A} & {N/A} & {48.0$\pm$23.6}  \\

{ ProtoContra \cite{protocontra}} & $\times$ & {99.5$\pm$25.5} & {72.6$\pm$16.3} & \textcolor{blue}{\textbf{67.9$\pm$18.4}} & {94.1$\pm$27.6} & {34.4$\pm$26.9} & {75.6$\pm$30.4} & {59.4$\pm$19.0}& {24.8$\pm$16.3} & {37.1$\pm$15.1} & {30.8$\pm$15.1}  & {N/A} & {N/A} & {45.8$\pm$18.7}  & {N/A} & {N/A} & {58.4$\pm$24.6}  \\

{ IAPC \cite{IAPC}} & $\times$ & {92.3$\pm$80.2} & \textcolor{blue}{\textbf{65.2$\pm$19.9}} & {69.8$\pm$21.2} & \textcolor{blue}{\textbf{82.0$\pm$37.9}} & {25.8$\pm$14.7} & {54.4$\pm$31.0} & {77.6$\pm$47.8}& {29.9$\pm$20.0} & {27.7$\pm$15.5} & {54.2$\pm$32.7}  & {N/A} & {N/A}& {45.3$\pm$25.0}  & \textcolor{blue}{\textbf{31.1$\pm$18.2}} & \textcolor{blue}{\textbf{31.2$\pm$19.7}} & {54.9$\pm$22.2} \\

{ DFG \cite{DFG}} & \checkmark & {98.5$\pm$21.0} & {77.0$\pm$21.1} & {72.5$\pm$9.4} & {N/A} & {10.7$\pm$8.6} & {67.3$\pm$24.5} & {54.0$\pm$16.6} & {20.5$\pm$6.3} & {30.5$\pm$7.9} & {30.5$\pm$15.5}   & {N/A} & {N/A}& {34.4$\pm$13.9} & {N/A} & {N/A} & {49.6$\pm$27.1} \\

{ IPLC \cite{iplc}} & \checkmark & {N/A} & {N/A} & {N/A} & {N/A} & {N/A} & \textcolor{blue}{\textbf{12.6$\pm$4.4}} & {40.9$\pm$13.5} & {50.9$\pm$16.3} & {52.8$\pm$11.3} & {39.6$\pm$11.3}  & {N/A} & {N/A} & {N/A}  & {N/A} & {N/A} & {39.4$\pm$14.4} \\ 

{ SRPL \cite{SRPL}} & \checkmark & {N/A} & {N/A} & {N/A} & {N/A} & \textcolor{blue}{\textbf{2.0$\pm$3.1}} & {31.2$\pm$30.0} & \textcolor{blue}{\textbf{15.3$\pm$18.2}} & \textcolor{blue}{\textbf{5.6$\pm$7.4}} & \textcolor{blue}{\textbf{20.5$\pm$7.8}} & \textcolor{blue}{\textbf{10.1$\pm$11.4}} & {N/A} & {N/A} & \textcolor{blue}{\textbf{29.1$\pm$24.8}}  & {N/A} & {N/A} & \textcolor{blue}{\textbf{16.3$\pm$10.4}} \\ \midrule

\textbf{Ours} & \checkmark & \textcolor{red}{\textbf{1.1$\pm$0.5}} & \textcolor{red}{\textbf{1.5$\pm$4.1}} & \textcolor{red}{\textbf{1.2$\pm$1.3}} & \textcolor{red}{\textbf{2.3$\pm$6.5}} & \textcolor{red}{\textbf{1.0$\pm$1.1}} & \textcolor{red}{\textbf{2.8$\pm$1.4}} & \textcolor{red}{\textbf{2.0$\pm$1.2}} & \textcolor{red}{\textbf{1.0$\pm$0.7}} & \textcolor{red}{\textbf{1.0$\pm$0.4}} & \textcolor{red}{\textbf{1.3$\pm$1.0}} & \textcolor{red}{\textbf{0.6$\pm$0.3}} & \textcolor{red}{\textbf{0.7$\pm$0.7}} & \textcolor{red}{\textbf{2.1$\pm$3.6}} & \textcolor{red}{\textbf{1.8$\pm$2.6}} & \textcolor{red}{\textbf{1.4$\pm$1.2}} & \textcolor{red}{\textbf{1.5$\pm$0.6}} \\

\bottomrule
\end{tabular}%
}

\resizebox{\columnwidth}{!}{%
\begin{tabular}{@{}cccccccccccccccc@{}}
\toprule
\multicolumn{15}{c}{\textbf{CT$\rightarrow$MR}} \\ \midrule

{ \textbf{Methods}} & { \textbf{VFM}} & { \textbf{Spleen}} & { \textbf{R. K}} & { \textbf{L. K}} & { \textbf{Gallb}} & { \textbf{Esoph}} & { \textbf{Liver}} & { \textbf{Stom}} & { \textbf{Aortic}} & { \textbf{V-Cava}} & { \textbf{Pancr}} & { \textbf{R. Adre}} & { \textbf{L. Adre}} & { \textbf{Duod}} & { \textbf{mASD}}\\ \midrule

{ Baseline} & $\times$ & {22.0$\pm$28.4} & {10.5$\pm$15.4} & {6.1$\pm$7.0} & {5.3$\pm$6.1} & 15.3$\pm$20.0 & {9.4$\pm$12.2} & 8.7$\pm$7.4 & 16.2$\pm$15.8 & {12.6$\pm$19.2} & 9.0$\pm$8.6 & 1.5$\pm$0.3 & 2.8$\pm$2.4  & 15.8$\pm$13.8 &  10.4$\pm$5.6\\

{ Supervised}  & $\times$& {1.0$\pm$0.1} & {1.3$\pm$1.4} & {0.9$\pm$0.4} & {1.3$\pm$1.3} & 1.9$\pm$1.6 & {2.2$\pm$0.9} & 1.8$\pm$0.8 & 1.3$\pm$1.1 & 0.9$\pm$0.4 & 1.5$\pm$0.5 & 1.1$\pm$0.6 & 1.4$\pm$1.0 & 3.1$\pm$1.7 &  1.5$\pm$0.6\\ \midrule

{ UPL \cite{upl}}  & $\times$ & {89.5$\pm$14.6} & \textcolor{blue}{\textbf{71.9$\pm$18.7}} & {62.9$\pm$25.4} & {N/A} & N/A & {53.4$\pm$20.8} & {61.1$\pm$25.5} & {14.3$\pm$19.7} & {N/A} & {33.3$\pm$24.8} & {N/A} & {N/A} & {N/A} &  55.2$\pm$23.0 \\

{ ProtoContra \cite{protocontra}} & $\times$ & {74.3$\pm$21.1} & {89.8$\pm$19.4} & {82.4$\pm$16.7} & {N/A} & {59.0$\pm$41.6} & {95.5$\pm$33.9} & {81.6$\pm$22.1} & {63.4$\pm$23.7} & {50.1$\pm$23.9} & {48.5$\pm$25.9} & \textcolor{blue}{\textbf{56.8$\pm$19.5}} & {N/A} & {N/A} &  70.1$\pm$16.0\\

{ IAPC \cite{IAPC}} & $\times$ & \textcolor{blue}{\textbf{71.5$\pm$48.1}} & {82.4$\pm$47.6} & {85.0$\pm$65.0} & {N/A} & 30.5$\pm$16.9 & 62.7$\pm$49.4 & 96.1$\pm$56.4 & 35.1$\pm$32.3 & \textcolor{blue}{\textbf{31.8$\pm$20.4}} & 74.8$\pm$39.1 & {N/A} & {N/A} & {N/A} &  63.3$\pm$23.5\\

{ DFG \cite{DFG}} & \checkmark & {99.9$\pm$27.0} & {83.8$\pm$3.3} & {80.9$\pm$17.0} & {N/A} & \textcolor{blue}{\textbf{18.0$\pm$17.3}} & {76.1$\pm$35.8} & 85.4$\pm$22.6 & 24.4$\pm$29.5 & 57.2$\pm$23.1 & 47.5$\pm$22.4 & {N/A} & {N/A} & {N/A} &  63.7$\pm$27.0 \\

{ IPLC \cite{iplc}} & \checkmark & {N/A} & {N/A} & {N/A} & {N/A}& {65.6$\pm$14.5} & \textcolor{blue}{\textbf{14.3$\pm$5.5}} & {31.9$\pm$12.0}  & 42.8$\pm$11.1 & 47.0$\pm$7.4 & 25.7$\pm$5.5 & {N/A} & {N/A} & 41.4$\pm$6.3 &  38.4$\pm$15.2\\

{ SRPL \cite{SRPL}} & \checkmark & {N/A} & {N/A} & \textcolor{blue}{\textbf{4.4$\pm$13.3}} & {N/A} & {N/A} & {16.9$\pm$22.7} & \textcolor{blue}{\textbf{8.1$\pm$9.3}} & \textcolor{blue}{\textbf{5.3$\pm$12.0}} & {N/A} & \textcolor{blue}{\textbf{10.5$\pm$11.5}} & {N/A} & {N/A} & \textcolor{blue}{\textbf{10.5$\pm$11.2}} &  \textcolor{blue}{\textbf{9.3$\pm$4.1}}\\ \midrule

{ \textbf{Ours}} & \checkmark & \textcolor{red}{\textbf{1.3$\pm$1.1}} & \textcolor{red}{\textbf{0.9$\pm$0.4}} & \textcolor{red}{\textbf{0.9$\pm$0.3}} & \textcolor{red}{\textbf{5.0$\pm$10.1}} & \textcolor{red}{\textbf{1.3$\pm$1.0}} & \textcolor{red}{\textbf{1.9$\pm$0.7}} & \textcolor{red}{\textbf{1.5$\pm$1.4}} & \textcolor{red}{\textbf{1.0$\pm$0.7}} & \textcolor{red}{\textbf{1.1$\pm$0.5}} & \textcolor{red}{\textbf{1.2$\pm$0.6}} & \textcolor{red}{\textbf{1.1$\pm$0.6}} & \textcolor{red}{\textbf{1.3$\pm$0.6}} & \textcolor{red}{\textbf{2.3$\pm$1.1}} &  \textcolor{red}{\textbf{1.6$\pm$1.1}}\\ 
\bottomrule
\end{tabular}%
}

\footnotesize{~~~~\textit{Note}:  An ASD of N/A indicates a complete prediction failure for the corresponding targets.}
\vspace{-3mm}
\end{table*}

As shown in \Cref{tab:AMOS_ASD_adaptation}, Tell2Adapt achieves consistently strong boundary accuracy across all abdominal organs. In MR$\rightarrow$CT, our framework attains a mean ASD of 1.5\,mm, representing a substantial reduction compared to Baseline and nearly matching Supervised of 1.4\,mm. 
This strong performance is further reflected in the per-organ results, where structurally diverse organs such as spleen (1.1\,mm), esophagus (1.0\,mm), and pancreas (1.3\,mm) exhibit highly accurate boundary localization.

The robustness of Tell2Adapt is further demonstrated in CT$\rightarrow$MR. Here, Tell2Adapt achieves a mean ASD of 1.6\,mm, again closely aligned with Supervised of 1.5\,mm. Notably, challenging small and elongated structures, including the right adrenal gland (1.1\,mm) and aorta (1.0\,mm) are segmented with high boundary precision.

Overall, the ASD results reinforce the findings derived from DICE. Tell2Adapt not only delivers accurate region-level segmentation but also preserves detailed boundary structure across all abdominal targets. This consistent performance across both MR$\rightarrow$CT and CT$\rightarrow$MR, on one of the most complex multi-organ abdominal benchmarks, highlights the strong generalization capability of Tell2Adapt and its effectiveness in overcoming severe domain shifts.
\subsection{Ablation Results in ASD for Abdominal Targets}

As stated in \Cref{sec:Ablation_Study}, we provide the corresponding ASD for our full ablation study on the abdominal targets. This complements DICE results presented in \Cref{tab:Ablation}, and the results detailed in \Cref{tab:Ablation_ASD} mirror and reinforce the trends observed for DICE.

Consistent with the DICE findings, CAPR emerges as the most influential method in improving boundary accuracy. Without CAPR, chaos prompts lead to a marked deterioration in boundary accuracy, with mean ASD increasing from 1.8\,mm to 2.8\,mm in MR$\rightarrow$CT and from 1.9\,mm to 3.5\,mm in CT$\rightarrow$MR. As illustrated by the comparison between Ours w/o CAPR (Chaotic Prompts) and Ours (Chaotic Prompts), several organs exhibit substantially larger surface distances, indicating pronounced boundary misalignment. This demonstrates CAPR’s ability to stabilize VFM guidance by reconstructing coherent and semantically aligned prompts, even when the original prompts are noisy and ambiguous.

Beyond its corrective role, CAPR also provides measurable optimization benefits. When comparing Ours (Normal Prompts) to Ours w/o CAPR (Normal Prompts), the mean ASD decreases notably from 2.4\,mm to 1.5\,mm in MR$\rightarrow$CT and from 2.8\,mm to 1.6\,mm in CT$\rightarrow$MR, demonstrating that CAPR improves boundary precision even when the initial prompts are already well-formed. This further confirms that CAPR not only rescues performance under chaos prompts but also improves semantic grounding under normal conditions. VPR provides an additional refinement by removing anatomically implausible components that can yield high ASD even when DICE appears reasonable. The increase in ASD from 1.5\,mm to 2.8\,mm in MR$\rightarrow$CT and from 1.6\,mm to 3.2\,mm in CT$\rightarrow$MR, when comparing Ours w/o VPR (Normal Prompts) to Ours (Normal Prompts) underscores the importance of incorporating anatomical priors. By eliminating false positives and enforcing shape plausibility, VPR enhances the boundary accuracy of Tell2Adapt.

Overall, the ASD ablation results validate the complementary effects of CAPR and VPR. CAPR provides robustness against prompt variability and enhances semantic alignment, while VPR enforces anatomical plausibility and boundary correctness. Their combined impact explains the substantial ASD reductions achieved by Tell2Adapt across both MR$\rightarrow$CT and CT$\rightarrow$MR, paralleling the improvements previously observed in DICE.

\begin{table*}[hb]
\centering
\caption{ASD (mm, mean $\pm$ std) of segmentation results in the ablation study on abdominal targets.}
\label{tab:Ablation_ASD}
\resizebox{\columnwidth}{!}{%
\begin{tabular}{@{}cccccccccccccccccc@{}}
\toprule
\multicolumn{17}{c}{\textbf{MR$\rightarrow$CT}} \\ \midrule
{ \textbf{Methods}} & { \textbf{Spleen}} & { \textbf{R. K}} & { \textbf{L. K}} & { \textbf{Gallb}} & { \textbf{Esoph}} & { \textbf{Liver}} & { \textbf{Stom}} & { \textbf{Aortic}} & { \textbf{V-Cava}} & { \textbf{Pancr}} & { \textbf{R. Adre}} & { \textbf{L. Adre}} & { \textbf{Duod}} & { \textbf{Bladder}} & { \textbf{Pros/Uter}} & { \textbf{mASD}}\\ \midrule
{ Baseline} & {17.5$\pm$19.2} & {17.2$\pm$18.5} & {15.5$\pm$17.8} & {23.1$\pm$30.7} & {7.8$\pm$9.6} & {11.4$\pm$7.2} & {15.3$\pm$14.6}& {8.9$\pm$9.9} & {6.9$\pm$11.2} & {9.1$\pm$10.7}  & {N/A} & {N/A} & {10.3$\pm$9.7}  & {N/A} & {N/A} & {13.0$\pm$4.8}  \\ \midrule
Ours w/o CAPR (Chaotic Prompts)& {1.7$\pm$1.1} & {N/A} & {2.5$\pm$0.5} & {2.6$\pm$1.4} & {2.6$\pm$0.8} & {N/A} & {N/A} & {2.4$\pm$0.7} & {5.4$\pm$9.3} & {N/A} & {1.7$\pm$0.4} & {N/A} & {3.9$\pm$2.4} & {N/A} & {2.6$\pm$2.5} & {2.8$\pm$1.1}\\
{ Ours w/o CAPR (Normal Prompts)} & {1.5$\pm$0.9} & \textcolor{blue}{\textbf{2.2$\pm$0.4}} & {2.1$\pm$1.5} & \textcolor{blue}{\textbf{2.6$\pm$0.8}} & {1.7$\pm$0.9} & {3.3$\pm$1.5} & {2.7$\pm$0.4} & {2.0$\pm$1.9} & {4.7$\pm$2.6} & {1.9$\pm$1.1} & {1.5$\pm$0.4} & {2.1$\pm$1.4} & \textcolor{blue}{\textbf{2.2$\pm$1.9}} & {2.9$\pm$1.1} & {2.4$\pm$0.7} & {2.4$\pm$8}\\
{ Ours w/o VPR (Normal Prompts)} & {2.1$\pm$1.5} & {2.4$\pm$1.3} & {2.4$\pm$2.1} & {2.9$\pm$1.8} & {2.5$\pm$0.7} & {3.5$\pm$3.0} & {3.7$\pm$2.9} & {2.2$\pm$1.7} & {5.3$\pm$4.8} & {2.6$\pm$1.8} & {1.5$\pm$1.2} & {2.4$\pm$0.9} & {3.1$\pm$1.7} & {3.4$\pm$2.7} & {2.6$\pm$3.3} & {2.8$\pm$0.9}\\ \midrule
{ Ours  (Chaotic Prompts) } & \textcolor{blue}{\textbf{1.2$\pm$0.4}} & {2.3$\pm$0.1} & \textcolor{blue}{\textbf{1.3$\pm$0.4}}  & {2.8$\pm$2.4} & \textcolor{blue}{\textbf{1.5$\pm$0.5}} & \textcolor{red}{\textbf{2.3$\pm$0.2}} & \textcolor{blue}{\textbf{2.1$\pm$5.5}} & \textcolor{blue}{\textbf{1.4$\pm$0.9}} & \textcolor{blue}{\textbf{1.5$\pm$0.2}} & \textcolor{red}{\textbf{0.9$\pm$0.4}} & \textcolor{blue}{\textbf{0.9$\pm$4.2}} & \textcolor{blue}{\textbf{1.0$\pm$4.4}} & {3.8$\pm$0.9} & \textcolor{red}{\textbf{1.7$\pm$2.0}} & \textcolor{blue}{\textbf{1.8$\pm$1.5}}  & \textcolor{blue}{\textbf{1.8$\pm$0.8}}\\ 
{ Ours (Normal Prompts)} & \textcolor{red}{\textbf{1.1$\pm$0.5}} & \textcolor{red}{\textbf{1.5$\pm$4.1}} & \textcolor{red}{\textbf{1.2$\pm$1.3}} & \textcolor{red}{\textbf{2.3$\pm$6.5}} & \textcolor{red}{\textbf{1.0$\pm$1.1}} & \textcolor{blue}{\textbf{2.8$\pm$1.4}} & \textcolor{red}{\textbf{2.0$\pm$1.2}} & \textcolor{red}{\textbf{1.0$\pm$0.7}} & \textcolor{red}{\textbf{1.0$\pm$0.4}} & \textcolor{blue}{\textbf{1.3$\pm$1.0}} & \textcolor{red}{\textbf{0.6$\pm$0.3}} & \textcolor{red}{\textbf{0.7$\pm$0.7}} & \textcolor{red}{\textbf{2.1$\pm$3.6}} & \textcolor{blue}{\textbf{1.8$\pm$2.6}} & \textcolor{red}{\textbf{1.4$\pm$1.2}} & \textcolor{red}{\textbf{1.5$\pm$0.6}}\\ 
\bottomrule
\end{tabular}%
}

\resizebox{\columnwidth}{!}{%
\begin{tabular}{@{}cccccccccccccccc@{}}
\toprule
\multicolumn{15}{c}{\textbf{CT$\rightarrow$MR}} \\ \midrule
{ \textbf{Methods}} & { \textbf{Spleen}} & { \textbf{R. K}} & { \textbf{L. K}} & { \textbf{Gallb}} & { \textbf{Esoph}} & { \textbf{Liver}} & { \textbf{Stom}} & { \textbf{Aortic}} & { \textbf{V-Cava}} & { \textbf{Pancr}} & { \textbf{R. Adre}} & { \textbf{L. Adre}} & { \textbf{Duod}} & { \textbf{mASD}}\\ \midrule
{ Baseline} & {22.0$\pm$28.4} & {10.5$\pm$15.4} & {6.1$\pm$7.0} & {5.3$\pm$6.1} & 15.3$\pm$20.0 & {9.4$\pm$12.2} & 8.7$\pm$7.4 & 16.2$\pm$15.8 & {12.6$\pm$19.2} & 9.0$\pm$8.6 & 1.5$\pm$0.3 & 2.8$\pm$2.4  & 15.8$\pm$13.8 &  10.4$\pm$5.6 \\ \midrule
{ Ours w/o CAPR (Chaotic Prompts)} & {2.7$\pm$1.1} & {N/A} & {2.5$\pm$1.3} & {5.6$\pm$0.7} & {2.7$\pm$0.3} & {N/A} & {N/A} & {2.4$\pm$0.7} & {5.4$\pm$9.3} & {N/A} & {2.7$\pm$0.9} & {N/A} & {3.8$\pm$1.4} & {3.5$\pm$1.2}\\
{ Ours w/o CAPR (Normal Prompts)} & {1.9$\pm$0.8} & {2.0$\pm$1.6} & {1.7$\pm$1.4} & {5.3$\pm$3.1} & \textcolor{blue}{\textbf{2.2$\pm$1.4}} & {2.5$\pm$0.4} & {2.7$\pm$2.1} & {2.0$\pm$1.8} & {4.7$\pm$3.5} & {3.1$\pm$2.4} & {2.5$\pm$1.4} & {2.4$\pm$1.1} & \textcolor{blue}{\textbf{2.9$\pm$1.5}} & {2.8$\pm$1.0}\\
{ Ours w/o VPR (Normal Prompts) } & {2.4$\pm$0.7} & {2.5$\pm$1.9} & {2.2$\pm$2.1} & {5.3$\pm$4.4} & {2.6$\pm$1.2} & {3.1$\pm$2.7} & {2.9$\pm$1.4} & {2.3$\pm$2.1} & {5.3$\pm$4.1} & {3.7$\pm$2.8} & {2.9$\pm$1.4} & {2.8$\pm$0.5} & {3.4$\pm$2.9} & {3.2$\pm$1.0}\\ \midrule
{ Ours (Chaotic Prompts) } & \textcolor{blue}{\textbf{1.4$\pm$0.2}} & \textcolor{blue}{\textbf{1.5$\pm$0.1}} & \textcolor{blue}{\textbf{0.9$\pm$0.4}} & \textcolor{red}{\textbf{4.6$\pm$1.8}} & {2.7$\pm$0.1} & \textcolor{red}{\textbf{1.0$\pm$0.0}} & \textcolor{blue}{\textbf{1.7$\pm$0.4}} & \textcolor{blue}{\textbf{1.4$\pm$0.5}} & \textcolor{blue}{\textbf{1.5$\pm$0.8}} & \textcolor{blue}{\textbf{1.6$\pm$0.4}} & \textcolor{blue}{\textbf{1.7$\pm$0.4}} & \textcolor{red}{\textbf{1.2$\pm$2.3}} & {3.8$\pm$1.4} & \textcolor{blue}{\textbf{1.9$\pm$1.1}}\\ 
{ Ours (Normal Prompts)} & \textcolor{red}{\textbf{1.3$\pm$1.1}} & \textcolor{red}{\textbf{0.9$\pm$0.4}} & \textcolor{red}{\textbf{0.9$\pm$0.3}} & \textcolor{blue}{\textbf{5.0$\pm$10.1}} & \textcolor{red}{\textbf{1.3$\pm$1.0}} & \textcolor{blue}{\textbf{1.9$\pm$0.7}} & \textcolor{red}{\textbf{1.5$\pm$1.4}} & \textcolor{red}{\textbf{1.0$\pm$0.7}} & \textcolor{red}{\textbf{1.1$\pm$0.5}} & \textcolor{red}{\textbf{1.2$\pm$0.6}} & \textcolor{red}{\textbf{1.1$\pm$0.6}} & \textcolor{blue}{\textbf{1.3$\pm$0.6}} & \textcolor{red}{\textbf{2.3$\pm$1.1}} &  \textcolor{red}{\textbf{1.6$\pm$1.1}}\\ 
\bottomrule
\end{tabular}%
}

\footnotesize{~~~~\textit{Note}:  An ASD of N/A indicates a complete prediction failure for the corresponding targets.}
\vspace{-3mm}
\end{table*}
\section{Details on Chaos Prompt in Ablation Study for CAPR}
\label{sec:appendix_B}
To strictly evaluate the robustness of CAPR against noisy real-world prompts, we introduced a systematic chaos prompt generation method. Unlike standard prompts, these prompts are intentionally corrupted with varying degrees of typographical errors and syntactic scrambling to simulate user input instability. This section details the mathematical formulation of these perturbations and provides the exact set of chaos prompts used in our ablation study to ensure the reproducibility of our robustness benchmarks.
\subsection{Chaos Prompts Generation Methodology}
To systematically evaluate the robustness of our method against noisy and corrupted text prompts, we introduce a controlled method for perturbing text prompts. This method generates prompt variations with a quantifiable chaos score to simulate real-world scenarios where prompts may contain typos, grammatical errors, or incomplete information. We inject noise through three primary categories of perturbations, with error rates dynamically scaled proportional to the target chaos level:
\begin{itemize}
    \item \textbf{Typographical Errors:} These simulate common spelling mistakes via character-level operations, including substitution (replacing a character with a random letter), transposition (swapping adjacent characters), and insertion (adding a random character), governed by the rate $r_{\text{spell}}$.
    \item \textbf{Syntactic Disruption:} We employ word order shuffling, where the positions of words within a prompt are randomly swapped to disrupt the grammatical structure, governed by the rate $r_{\text{shuffle}}$.
    \item \textbf{Information Loss:} We explicitly introduce character deletion, where characters are randomly removed from the text to simulate incomplete input, governed by the rate $r_{\text{remove}}$.
\end{itemize}

We quantify the chaos level using the normalized Levenshtein distance between the original and perturbed prompts:

\begin{equation}
S(P_{\text{pert}}) = \min\left(100, \frac{L(P_{\text{orig}}, P_{\text{pert}})}{\max(|P_{\text{orig}}|, |P_{\text{pert}}|)} \times 100\right)
\end{equation}

where $L(\cdot, \cdot)$ denotes the Levenshtein distance, $S(P_{\text{pert}})$ denotes the chaos score, $P_{\text{orig}}$ and $P_{\text{pert}}$ represent the original and perturbed prompts respectively, and $|\cdot|$ denotes the string length. The chaos score ranges from 0 (identical to original) to 100 (maximum perturbation). For each target chaos level $\tau \in \{5, 15, 30, 50, 75\}$, we employ an iterative optimization strategy. For each level, we generate 50 candidate versions and select the one whose actual chaos score is closest to the target:

\begin{equation}
P_{\tau}^* = \arg\min_{P \in \mathcal{C}_{\tau}} |S(P_{\text{pert}}) - \tau|
\end{equation}

where $\mathcal{C}_{\tau}$ is the set of candidate prompts generated for the target level $\tau$, and $S(P_{\text{pert}})$ computes the chaos score of prompt $P$. The perturbation rates for each operation are dynamically adjusted based on the target chaos level:
\begin{align}
r_{\text{spell}} &= 0.5 \times \frac{\tau}{100} \\
r_{\text{shuffle}} &= 0.7 \times \frac{\tau}{100} \\
r_{\text{remove}} &= 0.2 \times \frac{\tau}{100}
\end{align}
This method allows us to systematically evaluate model performance degradation across different levels of prompt corruption, ranging from minor typos (Chaos 5-15) to severely corrupted prompts (Chaos 50-75). For the ablation study presented in the main paper, we set the target chaos level to $\tau=75$ across all generated prompts, and the source code to generate chaos prompts is provided in the supplementary materials. The specific chaos prompts used in our experiments are explicitly listed in \Cref{tab:chaos_prompts}.
\subsection{Chaos Prompt in Ablation Study}
In this section, we provide a comprehensive list of perturbed text prompts used in the ablation study for CAPR. These chaos prompts were constructed to introduce a diverse range of textual noise to validate the necessity of CAPR. By documenting the exact strings in \Cref{tab:chaos_prompts}, researchers can replicate the specific noise conditions under which Tell2Adapt was evaluated. The table lists the chaos prompts used in both the MR$\rightarrow$CT and CT$\rightarrow$MR directions for all abdominal targets.
\begin{table*}[h]
\centering
\caption{Chaos prompt for ablation study in Tell2Adapt.}
\label{tab:chaos_prompts}
\footnotesize 
\begin{tabular}{@{}lp{5cm}p{5cm}@{}}
\toprule
\textbf{Target} & \textbf{Chaos Prompt in MR$\rightarrow$CT} & \textbf{Chaos Prompt in CT$\rightarrow$MR} \\ 
\midrule
Spleen & abdominal in CT Spleen. &  pSleen MR in abdomianl. \\
\addlinespace[0.2em]
Right Kidney & Right idney in abdominal CT. & MR kidey in Right abdominal. \\
\addlinespace[0.2em]
Left Kidney & Left abdominl kdne in CT. & Lewt kdiney in abdominl MR. \\
\addlinespace[0.2em]
Gallbladder & Gallbladedr in abdominal CT. & Gallbladder in abdomital MR. \\
\addlinespace[0.2em]
Esophagus & in CT abdominal Esophagus. & abdominal in MR Esophagus. \\
\addlinespace[0.2em]
Liver & Liiver in abdominal CT. & abdminal in MR Liver. \\
\addlinespace[0.2em]
Stomach & abdominal in CT Stomach. & in MR xbdominal Stomach. \\
\addlinespace[0.2em]
Aorta & in CT abdominal Aorta. & MR in Aorta abdominal. \\
\addlinespace[0.2em]
Vena-Cava & cava in abdominal CT Vedna. & Vena cav in abdominal MR. \\
\addlinespace[0.2em]
Pancreas & Pancreas CT. in bdominal. & Pancroeas in abdominal MR. \\
\addlinespace[0.2em]
Right Adrenal gland & glayd in adrenal Right abrdominal CT. & MR in Right adrenal abdomiial glandp. \\
\addlinespace[0.2em]
Left Adrenal gland & Left abdoimnal gland in adrenal CT. & abominal MR gland in adrenal Left. \\
\addlinespace[0.2em]
Duodenum & in bdominal Duodeunm CT. & Duodenum in abdominal MR. \\
\addlinespace[0.2em]
Bladder & dabdominal Bladdre in CT.& N/A \\
\addlinespace[0.2em]
Prostate/uterus & in Prostate/uterus CT abdominal. & N/A \\ 
\bottomrule
\end{tabular}
\end{table*}
\section{Meta-Prompt in CAPR and Full Prompts for All Targets}
\label{sec:appendix_C}
The Tell2Adapt's performance is guided by the quality of the prompts provided to the VFM. To ensure full transparency and allow for exact replication of our results, this section provides the meta-prompt in CAPR and the complete list of input prompts used for all anatomical targets evaluated in \Cref{sec:Experiments}. The prompts are organized by targets and domain adaptations in \Cref{tab:Abdominal_prompts}, \Cref{tab:Prompts_BraTS}, and \Cref{tab:Prompts_Cardiac-Polyp}.

\subsection{CAPR Meta-Prompt}
As detailed in \Cref{sec:CAPR} of the main paper, CAPR employs Qwen3-VL-8B-Instruct to normalize and canonicalize varied text prompts. The following listing provides the complete, unified meta-prompt given to the LLM. This single meta-prompt was used across all adaptation directions to parse, correct, and contextually enrich all prompts into the canonical format: [Target] in [Anatomical Site] [Modality].
\begin{lstlisting}[language=Python]
You are a meticulous assistant specializing in refining and standardizing biomedical image analysis prompts. Your user is preparing a list of text prompts for the BiomedParse vision foundation model, which requires precise, context-aware inputs.

Your task is to correct, refine, and normalize a list of "dirty" prompts I will provide.

Core Requirements
You must address two types of issues:

1. Linguistic Error Correction: You must find and fix all errors in each individual prompt, including but not limited to the following:
   - Spelling Mistakes
   - Letter Transposition
   - Missing Words/Characters
   - Repeated Words

2. Contextual Standardization: The BiomedParse model performs best when prompts are standardized. You must normalize every prompt to follow this strict format:

   [Target] in [Anatomical Site] [Modality]

To do this, you must first infer the global context (the default Modality and Site) from the entire list of prompts.

Step-by-Step Process
Before providing the final answer, you must perform the following internal reasoning steps:

Step 1: Analyze Global Context
Read the complete list of prompts I provide. Infer the most likely Modality and Anatomical Site that applies to the entire batch.

Step 2: State Your Context
I have analyzed the list. The inferred global context is:
Modality = [Your Inferred Modality]
Site = [Your Inferred Site]

Step 3: Iterate and Refine
Go through each prompt from the original list one by one.

- Analyze: The original prompt is: [original_prompt]
- Correct: The linguistic correction is: [corrected_prompt]
- Standardize: The prompt is missing context. Applying the global context, the final standardized prompt is: [standardized_prompt]
  (if the corrected prompt is liver and context is Abdomen CT, the final prompt is liver in abdomen CT)
- Exception: If a prompt already specifies its own context, like: tumor in brain MRI, respect that context and only correct its linguistic errors.

Final Output Format
Your final response to me MUST ONLY contain the list of fully corrected and standardized prompts.

- Do NOT include your internal Step-by-Step Thinking Process in the final output.
- Separate each standardized prompt with [SEP].

Example Output:
liver in abdomen CT[SEP]right kidney in abdomen CT[SEP]pancreas in abdomen CT[SEP]tumor in brain MRI

Task Starts Now
Here is the list of prompts for you to correct and normalize. Note the subprompt is also separated by [SEP]"""
\end{lstlisting}
\subsection{Target-Specific Input Prompts}
To facilitate full reproducibility of our extensive evaluations in \Cref{sec:Experiments}, we provide the exact prompts used for each anatomical target. These prompts were fed into Tell2Adapt to generate the quantitative and qualitative results presented in the main paper. The prompts are organized by target in \Cref{tab:Abdominal_prompts}, \Cref{tab:Prompts_BraTS}, and \Cref{tab:Prompts_Cardiac-Polyp}.
\begin{table*}[h]
\centering
\caption{Prompts for abdominal targets in Tell2Adapt.}
\label{tab:Abdominal_prompts}
\footnotesize 
\begin{tabular}{@{}lp{5cm}p{5cm}@{}}
\toprule
\textbf{Target} & \textbf{Prompt in MR$\rightarrow$CT} & \textbf{Prompt in CT$\rightarrow$MR} \\ 
\midrule
Spleen & Spleen in abdominal CT. & Spleen in abdominal MR. \\
\addlinespace[0.2em]
Right Kidney & Right kidney in abdominal CT. & Right kidney in abdominal MR. \\
\addlinespace[0.2em]
Left Kidney & Left kidney in abdominal CT. & Left kidney in abdominal MR. \\
\addlinespace[0.2em]
Gallbladder & Gallbladder in abdominal CT. & Gallbladder in abdominal MR. \\
\addlinespace[0.2em]
Esophagus & Esophagus in abdominal CT. & Esophagus in abdominal MR. \\
\addlinespace[0.2em]
Liver & Liver in abdominal CT. & Liver in abdominal MR. \\
\addlinespace[0.2em]
Stomach & Stomach in abdominal CT. & Stomach in abdominal MR. \\
\addlinespace[0.2em]
Aorta & Aorta in abdominal CT. & Aorta in abdominal MR. \\
\addlinespace[0.2em]
Vena-Cava & Vena-cava in abdominal CT. & Vena-cava in abdominal MR. \\
\addlinespace[0.2em]
Pancreas & Pancreas in abdominal CT. & Pancreas in abdominal MR. \\
\addlinespace[0.2em]
Right Adrenal gland & Right adrenal gland in abdominal CT. & Right adrenal gland in abdominal MR. \\
\addlinespace[0.2em]
Left Adrenal gland & Left adrenal gland in abdominal CT. & Left adrenal gland in abdominal MR. \\
\addlinespace[0.2em]
Duodenum & Duodenum in abdominal CT. & Duodenum in abdominal MR. \\
\addlinespace[0.2em]
Bladder & Bladder in abdominal CT.& N/A \\ 
\addlinespace[0.2em]
Prostate/uterus & Prostate/uterus in abdominal CT. & N/A \\ 
\bottomrule
\end{tabular}
\end{table*}
\begin{table*}[h]
\centering
\caption{Prompts for brain targets in Tell2Adapt.}
\label{tab:Prompts_BraTS}
\footnotesize 
\begin{tabular}{@{}lp{3.3cm}p{3.3cm}p{3.3cm}p{3.3cm}@{}}
\toprule
\textbf{Target} & \textbf{Prompt in T1n$\rightarrow$T2w} & \textbf{Prompt in T2w$\rightarrow$T1n} & \textbf{Prompt in T1c$\rightarrow$T2f} & \textbf{Prompt in T2f$\rightarrow$T1c} \\ 
\midrule
TC & Non-enhancing tumor core in head T2 weighted MR. & Non-enhancing tumor core in head MR naive T1. & Non-enhancing tumor core in head MR T2 FLAIR. & Non-enhancing tumor core in head post-contrast T1 MR. \\
\addlinespace[0.2em]
SNFH & Surrounding non-enhancing FLAIR hyperintensity in head T2 weighted MR & Surrounding non-enhancing FLAIR hyperintensity in head MR naive T1. & Surrounding non-enhancing FLAIR hyperintensity in head MR T2 FLAIR. & Surrounding non-enhancing FLAIR hyperintensity in head post-contrast T1 MR. \\
\addlinespace[0.2em]
ET & Enhancing tissue in head T2 weighted MR. & Enhancing tissue in head MR naive T1. & Enhancing tissue in head MR T2 FLAIR. & Enhancing tissue in head post-contrast T1 MR. \\
\addlinespace[0.2em]
RC & Resection cavity in head T2 weighted MR. & Resection cavity in head MR naive T1. & Resection cavity in head MR T2 FLAIR. & Resection cavity in head post-contrast T1 MR. \\
\bottomrule
\end{tabular}
\end{table*}
\begin{table*}[th]
\centering
\caption{Prompts for cardiac targets and polyp in Tell2Adapt.}
\label{tab:Prompts_Cardiac-Polyp}
\footnotesize
\begin{tabular}{@{}lp{5cm}p{5cm}@{}}
\toprule
\textbf{Target} & \textbf{Prompt in MR$\rightarrow$US} & \textbf{Prompt in US$\rightarrow$MR} \\ 
\midrule
Left Ventricle & Left ventricle in heart ultrasound. &  Left ventricle in cardiac MRI. \\
\addlinespace[0.2em]
Myocardium & Myocardium in heart ultrasound. & Myocardium in cardiac MRI. \\
\midrule
\textbf{Target} & \textbf{Prompt in Kvasir$\rightarrow$CVCDB} & \textbf{Prompt in CVCDB$\rightarrow$Kvasir} \\ 
\midrule
Polyp & Polyp in colon endoscopes. & Polyp in colon endoscopes. \\

\bottomrule
\end{tabular}
\end{table*}

\end{document}